\pdfoutput=1

\documentclass[11pt]{article}

\usepackage[]{acl}

\usepackage{times}
\usepackage{latexsym}

\usepackage[T1]{fontenc}

\usepackage[utf8]{inputenc}

\usepackage{microtype}

\usepackage{inconsolata}

\usepackage{graphicx}       
\usepackage{multirow}       
\usepackage{subcaption}     
\usepackage{bold-extra}     
\usepackage{bm}             
\usepackage[hang,flushmargin]{footmisc}  
\usepackage{stfloats}  
\usepackage{amsfonts,amsmath,amssymb}
\usepackage{pifont}
\usepackage{array,booktabs,makecell,tabularx}
\usepackage{lipsum}
\usepackage{enumitem}
\usepackage{xcolor}
\usepackage{csquotes}
\usepackage{makecell}
\usepackage{changepage}
\usepackage{tcolorbox}
\usepackage{mdframed}

\usepackage[]{todonotes}

\newcommand{\graph}{\textsc{NarCo} }
\newcommand{\graphns}{\textsc{NarCo}}

\newmdenv[
  leftline=true,
  rightline=false,
  topline=false,
  bottomline=false,
  linecolor=gray,
  linewidth=2pt,
  innerrightmargin=7pt,
  innerleftmargin=8,
  innertopmargin=3pt,
  innerbottommargin=3pt,
  leftmargin=6pt,
  rightmargin=5pt,
  skipabove=\medskipamount,
  skipbelow=\medskipamount
]{leftbarquote}

\usepackage{listings}
\lstdefinelanguage{prompt}{
    frame=l,
    framerule=3pt,
    framesep=8pt,
    basicstyle=\small\ttfamily,
    commentstyle=\color{cyan},
    morecomment=[l]{//},
    moredelim=[is][\color{brown}\bfseries]{<<<}{>>>},
    moredelim=[is][\color{magenta}\bfseries]{[[[}{]]]},
    moredelim=[is][\color{orange}\bfseries]{===}{===},
    moredelim=[is][\color{blue}\bfseries]{|||}{|||},
}

\title{Fine-Grained Modeling of Narrative Context:\\A Coherence Perspective via Retrospective Questions}

\author{Liyan Xu\quad\,\ Jiangnan Li\quad\,\ Mo Yu\thanks{Corresponding author.}\quad\,\ Jie Zhou\\
  Pattern Recognition Center, WeChat AI\\
  \normalsize \{{\texttt{liyanlxu,jiangnanli,withtomzhou}\}\texttt{@tencent.com}\quad\texttt{moyumyu@global.tencent.com}}}

\begin{document}
\maketitle

\begin{abstract}
This work introduces an original and practical paradigm for narrative comprehension, stemming from the characteristics that individual passages within narratives tend to be more cohesively related than isolated.
Complementary to the common end-to-end paradigm, we propose a fine-grained modeling of narrative context, by formulating a graph dubbed \graphns, which explicitly depicts task-agnostic coherence dependencies that are ready to be consumed by various downstream tasks. In particular, edges in \graph encompass free-form retrospective questions between context snippets, inspired by human cognitive perception that constantly reinstates relevant events from prior context. Importantly, our graph formalism is practically instantiated by LLMs without human annotations, through our designed two-stage prompting scheme.
To examine the graph properties and its utility, we conduct three studies in narratives, each from a unique angle: edge relation efficacy, local context enrichment, and broader application in QA. All tasks could benefit from the explicit coherence captured by \graphns.

\end{abstract}

\section{Introduction}
\label{sec:intro}

Since the advent of Large Language Models (LLMs), document comprehension has been improved significantly by simply employing the end-to-end generative paradigm.
Especially, with long context window enabled via techniques such as position interpolation \cite{long-llama,yarn}, cached or efficient attention \cite{long-mem,ge2024model,munkhdalai2024leave}, context compression or pruning \cite{chevalier-etal-2023-adapting,NEURIPS2023_cdaac2a0,ge2024incontext}, the end-to-end paradigm is deemed undoubtedly simple and effective for comprehension tasks (e.g. question answering) on various documents.

However, while the typical benchmarks are continually enhanced by more advanced LLMs \cite{llama2,jiang2024mixtral,openai2024gpt4}, the end-to-end paradigm may not suffice for all comprehension scenarios. In this work, we focus around the narrative context, i.e. stories or novels, and propose a conceptually original framework of fine-grained context modeling: a graph is formulated that depicts the relations between context snippets, abstracting over the context to reflect a high-level understanding of the narrative. The graph itself is practically realized by LLMs to harness their rapidly evolving strengths, and the resulting graph could serve to facilitate various downstream narrative comprehension tasks.

Our motivation arises from the distinctive nature of narratives: multiple development of characters or events  in a story could be entangled over long context ranges, where each local passage usually serves specific purposes for others. Thus, individual passages tend to be cohesively interconnected than being isolated.
As the end-to-end paradigm implicitly grasps these context connections through sequence modeling, our approach explicitly models these dependency relations to capture coherence, offering a directly-applicable alternative path orthogonal to the end-to-end paradigm.



Concretely, drawing inspiration from the cognitive process on narrative perception, whereas humans can constantly reinstate relevant or causal events from past context during reading \cite{backward1,backward2}, our formalism, termed NARrative COgnition graph (\graphns), splits the entire context into chunks that act as graph nodes, with edges representing the relations between node pairs.
In particular, edge relations are constituted by free-form questions. As humans could relate to past context in retrospect, accordingly, each question in \graph edges arises from the succeeding node (\textit{latter} context), asking necessary background or causes that can be clarified by the preceding node (\textit{prior} context). Hence, graph edges consist of inquisitive questions that naturally reflect retrospection. Overall, the resulting graph explicitly depicts task-agnostic understanding of fine-grained coherence flow that could be flexibly utilized by downstream tasks.

Though our graph formulation partially shares motivations with discourse parsing that characterizes how each proposition relates to others within a close context \cite{discourse}, our method targets on practical utility for narrative comprehension, where edges in \graph are designed to be easily obtained and effectively consumed by downstream tasks.
Consequently, \graph is formulated in a different scope from discourse parsing by two main perspectives.
First, as most discourse frameworks, such as Rhetorical Structure Theory \cite{rst}, Penn Discourse Treebank \cite{pdtb}, or the recent Questions Under Discussion (QUD) \cite{ko-etal-2022-discourse,ko-etal-2023-discourse} are rooted upon linguistic principles, their relation types are oriented for formal discourse analysis, requiring trained experts to annotate edges according to a defined linguistic taxonomy. Whilst for \graphns, the relation space is larger without taxonomy constraints, offering diverse high-level semantic signals for narrative tasks. 
Second, \graph practically leverages LLMs to derive edge relations, without reliance on human annotations. Thus, the edge quality is not restricted by annotation resources, and shall be continuously enhanced along with the ongoing LLM advancement.

The key difficulty of \graph lies in the edge realization between two nodes, which itself demands strong context understanding to determine which aspects to inquire upon the context, and to assess their saliency for comprehension. Such process is especially strenuous due to the large hypothesis space compared to conventional discourse formalisms.
To this end, we pose soft semantic constraints on relations, and employ LLM's capabilities to construct edges automatically through our proposed prompting scheme, of which consists a question generation stage and a self verification stage (Section~\ref{sec:graph}).
The obtained edges could then be utilized by downstream tasks in two primary ways.
First, edges themselves directly provide information flow to guide various comprehension tasks.
Second, they offer global coherence view for each node, thereby augmenting the local context to deepen the digest of independent passages.




To empirically demonstrate the practical utility of \graphns, we present three studies on narrative comprehension tasks, each from a unique angle:

\noindent $\bullet$ Our first study examines the \textbf{edge efficacy} on \emph{whether the relation questions capture capable retrospective coherence} (Section~\ref{sec:task-recap}). We conduct experiments on the recap identification task \cite{recap}, where \graph is shown to recognize coherence dependencies between context, boosting up to 4.7 F1 over the GPT-4 baseline.

\noindent $\bullet$ Our second study concerns the exploitation of \textbf{enriched local embeddings}, by \emph{injecting edges of relation dependencies into node representation} (Section~\ref{sec:task-rt}). Evaluated on the plot retrieval task \cite{xu2023plot}, our proposed approach with \graph outperforms the zero-shot baseline by 3\% and the supervised baseline by 2.2\%.

\noindent $\bullet$ Lastly, we utilize \graph in a long document question answering task (Section~\ref{sec:task-qa}), as a broader application of \textbf{Retrieval-Augmented Generation} (RAG) \cite{rag}. Experiments on QuALITY that requires global context evidence \cite{pang-etal-2022-quality} suggest that, \graph consistently raises zero-shot accuracy by 2-5\% upon retrieval-based baselines with various LLMs, able to identify more relevant context through edge relations.

Overall, our key contributions in this work are:
\begin{itemize}[noitemsep,nolistsep,leftmargin=*]
    \item We propose a new paradigm of fine-grained context modeling to facilitate narrative comprehension, orthogonal to the end-to-end paradigm.
    \item Our introduced \graph framework describes flexible relations of context dependencies by retrospective questions, which are realized by LLMs through our designed prompting scheme, without reliance on human annotations.
    \item We present three studies effectively utilizing \graph on narrative tasks, empirically examining its edge properties and broader utilization.
\end{itemize}

\section{Related Work}
\label{sec:related}

\begin{figure*}[th]
\centering
\includegraphics[width=\textwidth]{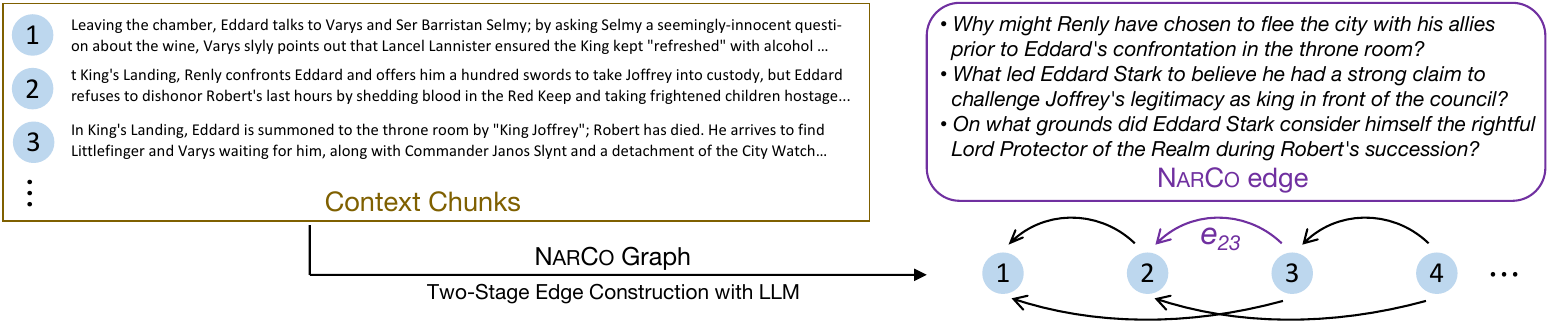}
\caption{Our proposed \graph graph described in Section~\ref{sec:graph}, with retrospective questions connecting two nodes.}
\label{fig:graph}
\vspace{-2ex}
\end{figure*}

\paragraph{Questions Under Discussion}

QUD is a linguistic framework with rich history that approaches discourse and pragmatics analysis by repeatedly resolving queries triggered by prior context \cite{qud_1995,qud_1996,qud-survey}. 
QUD has been adapted by recent works for discourse analysis \cite{de-kuthy-etal-2018-qud,de-kuthy-etal-2020-towards,ko-etal-2020-inquisitive,ko-etal-2022-discourse,ko-etal-2023-discourse} or other applications \cite{wu-etal-2023-elaborative,decontextualization-newman-2023}. Our proposed \graph also adopts question-form relations; though, the scope and motivation is different from discourse analysis. Consequently, \graph differs from QUD works considerably on the following design choices.

• \textbf{Coarse Granularity} \;
While QUD tends to employ sentences as the basic discourse unit, \graph opts for a coarser granularity, adopting passages (or chunks) as the graph nodes. It is driven by the fact that in narratives, complex events or interactions may often be conveyed beyond sentence-level, thus relations in \graph could target higher-level understanding between context.

• \textbf{Retrospection-Oriented} \;
Unlike conventional QUD that inquires from prior context to be addressed by subsequent context (forward direction), which could yield unanswerable questions \cite{westera-etal-2020-ted,ko-etal-2020-inquisitive}, \graph takes the \emph{backward} direction, by asking retrospective questions from latter context, such that all generated questions in \graph are naturally grounded by the corresponding prior context.

• \textbf{Precision-Focused} \;
Unlike previous QUD works that require dedicated human annotations, \graph is formulated attainable by LLMs. Accordingly, we prioritize precision over recall for practical instantiation of graph edges, and do not necessitate strict linguistic criteria, as long as edges contribute positively for narrative understanding.

\paragraph{Narrative Comprehension Assessments}
A major task direction on narratives is question answering (QA), where past works have proposed several datasets with human annotations, such as NarrativeQA \cite{kocisky-etal-2018-narrativeqa}, TellMeWhy \cite{lal-etal-2021-tellmewhy}, \textsc{FairytaleQA} \cite{xu-etal-2022-fantastic}, QuALITY \cite{pang-etal-2022-quality}. We adopt QuALITY as the broader application in this work, due to its challenging long context, requirement of global evidences, and simple evaluation by multi-choices.

Recently, several tasks have emerged focusing on modeling the reading process of long narratives, including TVShowGuess \cite{sang-etal-2022-tvshowguess}, \textsc{PersoNet} \cite{yu-etal-2023-personality}, \textsc{ToM-in-AMC} \cite{tominamc}, and retrieval tasks such as RELiC \cite{thai-etal-2022-relic}, \textsc{PlotRetrieval} \cite{xu2023plot}. These tasks require a holistic understanding of the long narratives to enhance contextual comprehension of specific segments. We reckon the significance of explicitly modeling context dependencies as a crucial aspect of narrative comprehension, motivating the inception of .

\paragraph{LLM Understanding and Reasoning}
LLMs have demonstrated remarkable capabilities on a wide spectrum of comprehension and reasoning tasks \cite{chen2024comm,sun2024itd}. The simple end-to-end solution is especially appealing with long context window enabled, using techniques such as scaling positional embeddings \cite{chen2023extending,long-llama,yarn}, efficient attention \cite{munkhdalai2024leave}, cached attention \cite{long-mem,ge2024model}, recurrent attention \cite{dai-etal-2019-transformer}, context compression \cite{chevalier-etal-2023-adapting,ge2024incontext}, context pruning \cite{NEURIPS2023_cdaac2a0}, etc. Though being effective, certain narrative tasks demand beyond the end-to-end solution. Recently, new methods have been proposed for fine-grained task processing, e.g. reading agents such as \textsc{MemWalker} \cite{mem-walk}. Nevertheless, our proposed approach depicts explicit context dependencies as an alternative paradigm, which is orthogonal to the existing directions and could be even further combined.

\paragraph{Structured Representation}
Various relational structures in text documents has attracted much attention by previous works, such as syntactic relations \cite{strubell-etal-2018-linguistically,Xu_isdg_2022}, discourse relations \cite{ji-smith-2017-neural,nair-etal-2023-drilling,hu-wan-2023-exploring}, entity or event relations \cite{ding2019cognitive,li2020connecting,li2021timeline,xu-choi-2020-revealing,xu-choi-2022-modeling,nguyen-etal-2022-joint}.
As all these structures encompass pre-defined taxonomies on edge types, our propose graph representation is motivated to comprise open-world edge types that have been practiced in other tasks \cite{wu-etal-2019-open,xu-etal-2023-towards,sig}, while being practical and attainable by LLMs without requiring efforts of human annotations.

\section{\graphns: Narrative Cognition Graph}
\label{sec:graph}

In this section, we start by delineating our graph formulation, which is itself not tied to any particular implementation. Subsequently, we elaborate our specific graph realization using LLMs, without dependence on human annotations.

\subsection{Graph Formulation}
\label{ssec:graph-formulation}

\paragraph{Nodes}
For a narrative, the entire context is split into short consecutive chunks (or passages), such that each is within a maximum word limit and constituted by sentences or paragraphs. Graph nodes are then all the chunks adhering the left-to-right sequential order, denoted by $\mathcal{V} = \{v_1, v_2, .., v_N\}$,  with $N$ being the total number of chunks.

\paragraph{Edges}
An edge connecting two nodes indicates the relations between the context. These relations are articulated as free-form questions that are not constrained by fixed taxonomies. All edges follow the backward direction, such that for an edge $e_{ij}$ ($i < j$), the expressed questions always arise from the succeeding node $v_j$, asking clarification regarding specific events or situations appeared in $v_j$, which could be addressed by the preceding context $v_i$. 
Since the hypothesis space is huge without any regularization, we pose soft semantic constraints on questions, such that questions should primarily reflect on causal and temporal relations, which are significant to the narrative coherence.

Functionally speaking, these backward edges resemble the human cognitive process for narrative perception: when reading a certain passage, humans are able to reinstate previous relevant parts in retrospect that lay out the build-up or causes, so to achieve a coherent comprehension of the global context \cite{backward1,backward2,backward3}. Unlike conventional QUD that features curiosity-driven questions in a forward direction, which could yield unanswerable questions, all edge in \graph are fully grounded by the prior context, since all retrospective questions are addressable by prior nodes.

Derived upon the above formulation, an edge $e_{ij}$ in \graph has the following features:
\begin{itemize}[noitemsep,nolistsep,leftmargin=*]
    \item It may have zero or many questions. An empty edge (zero questions) signifies $v_j$ is vaguely independent from $v_i$ in terms of coherence.
    \item Each question should be salient towards the comprehension of narrative development, rather than inquiring trivial details. Hence, the number of questions in an edge should reflect how cohesively related between two nodes.
    \item As we adopt coarse granularity for nodes, questions could probe higher-level relations based on the extrapolation over multiple sentences, which may be useful towards broader understanding.
\end{itemize}

\subsection{Graph Realization}
\label{ssec:graph-realization}

To obtain graph nodes, the full context is split by paragraph and sentence boundaries. We impose each node within 240 words in this work, though the exact limit can be task-specific.
For a graph characterized by $N$ total nodes, there are $O(N^2)$ full edges available, which can become cumbersome and excessive. It is also task-dependent to determine which pairs of nodes should be gathered edges upon, e.g. for enriching local representation, it is sufficient to obtain relation dependencies from neighboring nodes within a context window.

Despite the daunting formulation on edge relations, the emergence of LLMs presents an opportunity: through designed LLM prompting, it becomes conceivable to actualize the entire graph without any human annotations involved. To this end, we introduce a two-stage prompting scheme to tackle the challenging edge construction.

\paragraph{Question Generation}
For an edge $e_{ij}$ to be instantiated, LLMs need to determine important aspects to ask upon $v_j$ that reflect the retrospective coherence towards the prior context in $v_i$. Similar utilization of LLMs for question generation (QG) has been explored in other applications, such as performing QG for QUD \cite{wu-etal-2023-qudeval} and passage decontextualization \cite{decontextualization-newman-2023}, where a LLM is prompted to generate questions directly based on task-specific criteria.
For our case, such direct generation can be briefly outlined as:

\begin{leftbarquote}
\tt
\small
Given a current context $v_j$ and its prior context $v_i$, generate questions upon $v_j$, such that each question asks about the cause or background of specific events or situations in $v_j$, which can be clarified by $v_i$, so to reflect their causal or temporal relations between the two context.
\end{leftbarquote}

\noindent However, our preliminary experiments suggest that although LLMs can generate plausible questions by following the instructions, their quality is often unsatisfactory for \graphns, with common errors as follows (examples in Appx~\ref{appx:examples}): 
\begin{itemize}[noitemsep,nolistsep,leftmargin=*]
    \item \textit{Self-answerable}: LLMs often ask questions upon $v_j$ but also answerable by $v_j$ as well. Such self-answerable pattern aligns with the more conventional QG setting \cite{du-etal-2017-learning} that may exist plentifully during the supervised finetuning of LLMs, causing a bias towards this type of question. However, they are not desirable for \graphns, since they do not express dependencies between nodes to reflect their relations.
    \item \textit{Hallucination}: LLMs could hallucinate the relations of two nodes by guessing and inferring extra underlying connections not grounded by the provided context, resulting in questions not directly answerable by $v_i$.
\end{itemize}

\noindent In essence, QG for \graph requires LLM simultaneously aware of questions being: 1) arising from $v_j$; 2) not answerable by $v_j$; 3) answerable by $v_i$. As this is empirically challenging even for strong LLMs (e.g. GPT-4), we perform QG with two heuristic turns that can be viewed as human-guided Chain-of-Thoughts \cite{cot}:

\begin{leftbarquote}
\tt
\small
1.\;\;List concrete parts in $v_i$ that contribute as the preceding background or cause for specific events or situations mentioned in $v_j$, along with brief explanations.\\

\noindent2.\;\;Convert each above listed connection to a question, such that it asks about the cause or background upon $v_j$ and can be clarified by the corresponding concrete part in $v_i$, helpful to comprehend their causal or temporal relations.
\end{leftbarquote}

\noindent The designed two-turn QG scheme yields higher-quality questions than the rudimentary generation, mainly alleviating the self-answerable problem. However, noisy questions of the two identified error types still occur due to imperfect instruction following by LLMs. In light of these noises, we apply an optional second stage to filter out noisy questions through self verification.

\paragraph{Self Verification}
The second stage takes the generated questions from QG and in turn, performs question answering on the context:

\begin{leftbarquote}
\tt
\small
Given a context $\mathcal{C}_{ij}$ and a related question, determine whether it is answerable. If yes, reason the answer and provide original sentences of key supporting evidences.
\end{leftbarquote}

\noindent In Particular, $\mathcal{C}_{ij}$ is the concatenated context from $v_i$ and $v_j$ without disclosing their boundary. If the question is answerable, we then parse the response and identify whether the supporting sentences are from the prior context $v_i$. If not, the question is attested noisy and gets discarded, as it does not bridge the two context effectively.

With the second stage, only questions that could be answered by prior nodes are eventually retained in \graphns, being a precision-focused approach. In this work, we adopt GPT-4 for the challenging QG stage, and ChatGPT for the easier verification stage. \graph may also be derived with strong open-source LLMs as well. Our full prompts and more details are provided in Appx~\ref{appx:prompts}.


As \graph targets the practical utility to facilitate narrative comprehension, the obtained edges shall be directly consumed by downstream tasks.
Sections~\ref{sec:task-recap}-\ref{sec:task-qa} present three empirical studies, each from a distinctive perspective, to examine the edge properties and their utilization.


\begin{figure}[ht!]
\centering
\includegraphics[width=\columnwidth]{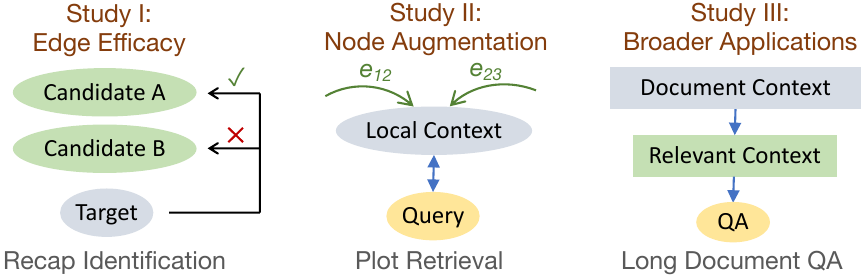}
\caption{Three presented studies leveraging \graphns.}
\label{fig:studies}
\vspace{-1ex}
\end{figure}

\section{Study I: Edge Efficacy}
\label{sec:task-recap}

Our first study examines the graph edges on whether they express useful relations, such that the generated retrospective questions should bridge the coherence between two context. For appropriate assessment, we conduct recap identification on RECIDENT dataset \cite{recap}, a task on narratives that identifies whether certain preceding snippets can function as a recap or prelude to the audience in regards to a current context.

Concretely, the input takes a short snippet from a novel or show script, along with a provided list of its preceding snippets. The task resolves which preceding snippets are directly related with the current one in terms of plot progression, requiring contextual understanding of narrative development.
As \graph is designed to capture the coherence relations between context, edges of retrospective questions could be leveraged to link the current snippet to related preceding ones. Therefore, RECIDENT serves as a natural testbed for comprehensive evaluation of edge efficacy.

\subsection{Approach}
\label{ssec:recap-approach}

For this study, our proposed approach targets upon the zero-shot baseline with LLMs in \cite{recap}, where ChatGPT is originally asked to select the related recap snippets from the list of preceding candidates based on their text content.

With \graphns, we regard each current snippet as a \textbf{t}arget graph node $v_t$, and the list of its $N$ preceding snippets $\{v_c | c=1,..,N\}$ as the \textbf{c}andidates. For $v_t$ and each of its candidate $v_c$, the edge is realized as $e_{ct}$.
As questions in $e_{ct}$ should reflect causal or temporal coherence, we directly utilize these questions in two following ways.

\paragraph{Edge Relations}
Normally, candidate snippets are processed by their text content as in the baseline. To evaluate the coherence depicted by edges, we instead propose to identify recap snippets solely based on the edge relations: for a candidate node $v_c$, we concatenate all its questions in edge $e_{ct}$, denoted by $q_c$, to identify recap, and completely neglect original text content, so to ensure an entirely isolated assessment of edge relations.

Specifically, given the context of a target snippet $v_t$, and $N$ candidates $\{q_c | c=1,..,N\}$ represented by questions, we now ask a LLM to score which $q_c$ addresses important questions that are significant to comprehend the current context, with higher scores indicating better overall questions that provide recap information. Candidates with empty edges are directly assigned 0 score.

\paragraph{Edge Degrees}
Alternatively, as mentioned in Section~\ref{ssec:graph-formulation}, the number of questions between two nodes could suggest how cohesively related they are.
We take this number as the edge degree, and propose to simply deem it as the score to rank candidates, without any inference on the node context or edge relations at all. Though being rather unconventional, ranking candidates by edge degrees further reflects the edge quality.

With either the relation score or degree score, it can be used standalone or interpolated with the baseline selection. More formally, we obtain the rank $\in [1, N]$ of each candidate $i$ by relation scores, denoted as $r^{rel}_i$, and the rank by degree scores $r^{deg}_i$, along with binary selection $b_i$ from the original baseline. The final score $s$ of each candidate is:
\begin{align}
\label{eq:recap}
    s_i = \alpha \cdot r^{rel}_i + \beta \cdot r^{deg}_i - \lambda \cdot \mathbb{I}(b_i)
\end{align}
$\mathbb{I}$ is the indicator function that boosts the baseline decision $b_i$ by $\lambda$ rank; relation and degree ranks are interpolated by $\alpha$ and $\beta$. The final score is then ranked to select top candidates with recap information (lower is better). Setting $\alpha$/$\beta$/$\lambda$ to 0 can thereby evaluate each method standalone.

\subsection{Experiments}
\label{ssec:recap-exp}

\paragraph{Data} As RECIDENT includes multiple novels and show scripts, we pick one classic novel \textit{Notre-Dame de Paris} (NDP) in English and one TV show \textit{Game of Thrones} (GOT) to reduce the evaluation API cost from OpenAI. The test set of each source consists of 169 / 204 target snippets respectively. Each target is provided 60 candidate snippets, with 5.6 / 4.9 candidates being positive on average.

\paragraph{Evaluation Metric}
We follow \citet{recap} and adopt F1@5 (F1 on top-5 selected candidates) as the main evaluation metric.

\paragraph{Methods}
We conduct zero-shot LLM experiments with both ChatGPT (\textit{gpt-3.5-turbo-1106}) and GPT-4 (\textit{gpt-4-1106-preview}) from OpenAI.
\begin{itemize}[noitemsep,nolistsep,leftmargin=*]
    \item \textbf{\texttt{BL}}: the original ChatGPT baseline (\textit{Listwise + Char-Filter} from \citet{recap}.) We additionally run GPT-4 for comprehensive evaluation.
    \item \textbf{\texttt{Rel}}: standalone ranking by edge relations, without using any candidate context itself.
    \item \textbf{\texttt{Full}}: full interpolation by Eq~\eqref{eq:recap} with both edge relations and degrees. Coefficients are set through a holdout set from another novel.
\end{itemize}

\begin{table}[htp!]
\centering
\resizebox{\columnwidth}{!}{
\begin{tabular}{l|ccccccc}
\toprule
& \multicolumn{3}{c}{NDP} && \multicolumn{3}{c}{GOT} \\
\cmidrule{2-4} \cmidrule{6-8}
& P@5 & R@5 & F@5 && P@5 & R@5 & F@5 \\
\midrule
& \multicolumn{7}{c}{\it ChatGPT} \\
\texttt{BL} & 22.22 & 22.97 & 22.59 && 31.94 & 38.87 & 35.07 \\
\texttt{Rel} & 22.84 & 23.34 & 23.09 && 28.63 & 37.09 & 32.31 \\
\texttt{Full} & 26.86 & 28.16 & \bf 27.50 && 33.04 & 43.27 & \bf 37.47 \\
\midrule
& \multicolumn{7}{c}{\it GPT-4} \\
\texttt{BL} & 25.34 & 25.53 & 25.44 && 31.49 & 40.38 & 35.38 \\
\texttt{Rel} & 26.39 & 27.23 & 26.80 && 31.18 & 42.05 & 35.81 \\
\texttt{Full} & 29.11 & 28.74 & \bf 28.92 && 34.90 & 46.93 & \bf 40.03 \\
\bottomrule
\end{tabular}}
\caption{Zero-shot evaluation on the test set of RECIDENT for recap identification (Section~\ref{ssec:recap-exp}). Our approaches with \graph achieve significant improvement upon the baseline (\texttt{BL}) for both ChatGPT and GPT-4.}
\label{tab:recap-result}
\vspace{-1ex}
\end{table}

\subsection{Results}
\label{ssec:recap-result}

Table~\ref{tab:recap-result} shows the zero-shot evaluation results on the test set of RECIDENT.
Notably, the interpolation with \graph edges (\texttt{Full}) consistently brings improvement upon the baseline (\texttt{BL}), by 4.9 / 2.4 F1 on NDP / GOT respectively with ChatGPT, up to a 21.7\% relative improvement. The stronger GPT-4 boosts performance for all methods as expected, and \graph still advances 3.5 / 4.7 F1 upon \texttt{BL} on NDP / GOT as well.

Moreover, selection solely based on edge relations  (\texttt{Rel}) obtains comparable or better performance than the baseline, with the only exception on GOT with ChatGPT. Overall, Table~\ref{tab:recap-result} effectively demonstrates the edge efficacy of \graph that expresses coherence through retrospective questions.

For in-depth analysis, we further perform two additional evaluation with ChatGPT:
\begin{itemize}[noitemsep,nolistsep,leftmargin=*]
    \item \textbf{\texttt{Deg}}: standalone ranking by edge degrees; for tied degrees, closer candidates are prioritized.
    \item \textbf{\texttt{Full\textsuperscript{-F}}}: the \texttt{Full} setting with all generated questions, without \textbf{F}iltering by self verification.
\end{itemize}

\begin{table}[htbp!]
\centering
\resizebox{\columnwidth}{!}{
\begin{tabular}{l|ccccccc}
\toprule
& \multicolumn{3}{c}{NDP} && \multicolumn{3}{c}{GOT} \\
\cmidrule{2-4} \cmidrule{6-8}
& P@5 & R@5 & F@5 && P@5 & R@5 & F@5 \\
\midrule
\texttt{BL} & 22.22 & 22.97 & 22.59 && 31.94 & 38.87 & 35.07 \\
\texttt{Full} & 26.86 & 28.16 & \bf 27.50 && 33.04 & 43.27 & \bf 37.47 \\
\midrule
\midrule
\texttt{Deg} & 23.31 & 24.44 & 23.86 && 27.45 & 37.67 & 31.76\\
\texttt{Full\textsuperscript{-F}} & 26.39 & 27.06 & 26.72 && 33.24 & 42.57 & 37.33 \\
\bottomrule
\end{tabular}}
\caption{Zero-shot evaluation with ChatGPT, using \graph edge degrees (\texttt{Deg}) and all questions (\texttt{Full\textsuperscript{-F}}).}
\label{tab:recap-result2}
\vspace{-1ex}
\end{table}

Table~\ref{tab:recap-result2} shows the additional evaluation results, where ranking by edge degrees of \graph exhibits decent performance. It even surpasses the baseline on NDP by 1+\%, which is impressive for the fact that it does not undergo any task-specific inference. Understandably, it indeed lags behind the baseline on GOT by a noticeable margin. 

For \texttt{Full\textsuperscript{-F}}, the degradation is trivial from \texttt{Full}. It is also expected, as the LLM scoring on relations is based on the presence of ``good'' questions that reflect recap information, which should be retained by the verification stage. Thus, our approach with \graph is shown robust against noisy questions.

\subsection{Graph Insights}

The majority of generated questions in \graph are \textit{what}/\textit{why}/\textit{how}-type of questions.
Their ratios are provided in Table~\ref{tab:graph-stats}, along with the averaged number of questions per edge before / after the self verification stage (Section~\ref{ssec:graph-realization}).

\begin{table}[htbp!]
\centering
\resizebox{0.8\columnwidth}{!}{
\begin{tabular}{l|cc}
\toprule
& NDP & GOT \\
\midrule
\textit{What}-Questions Ratio & 61.5\% & 58.4\% \\
\textit{Why}-Questions Ratio & 26.5\% & 25.2\% \\
\textit{How}-Questions Ratio & 7.8\% & 14.0\% \\
\midrule
\# Questions per Edge & 3.4 & 3.5 \\
\; + Self Verification & 1.9 & 2.0 \\
\bottomrule
\end{tabular}}
\caption{Statistics of \graph in Study I (Section~\ref{sec:task-recap}).}
\label{tab:graph-stats}
\vspace{-1ex}
\end{table}

\section{Study II: Node Augmentation}
\label{sec:task-rt}

Our second study underscores the \graph utility of local context augmentation, examining whether the graph typology could enrich node representation with global contextual information. 

Specifically, for a node $v_j$, a preceding node $v_i$ and succeeding node $v_k$ such that $i < j < k$, $e_{ij}$ depicts \textit{outgoing} questions arising from $v_j$ to $v_i$, and $e_{jk}$ specifies \textit{incoming} questions from $v_k$ that can be clarified by $e_j$. These questions either highlight important aspects of events or situations in the current context, or provide implication of subsequent development. Such auxiliary information from neighboring nodes is especially useful for retrieval on narratives, as each passage tends to be more interconnected with others than isolated.

We hence investigate if an embedding function on top of \graph could lead to enriched local representation.
Towards this objective, we consider the plot retrieval task defined in \cite{xu2023plot}, which aims to find the most relevant story snippets given a query of short plot description.
It is challenging as queries are often abstract based on readers' overall understanding of the stories, requiring essential background information clarified on candidates, analogous to the concept of \textit{decontexualization} \cite{decontextualization-choi-2021}. Retrieval on narratives thereby fits our evaluation purpose well.

\subsection{Approach}
\label{ssec:rt-approach}

For this task, candidate snippets from stories are retrieved upon a given query. We regard all candidate snippets as graph nodes to be retrieved from, and derive \graph edges of neighboring nodes. Our proposed method focuses on fusing edge questions into node representation for enhanced retrieval.

\citet{xu2023plot} follows the classic paradigm of contrastive learning that learns a BERT-based encoder \cite{bert} on queries and candidates. As its trained model is not released as of this writing, our approach adopts the public BGE encoder \cite{bge} in this work that ranks top on the MTEB leaderboard\footnote{\url{https://huggingface.co/spaces/mteb/leaderboard}}. For comprehensive evaluation, we propose methods with \graph for both zero-shot and supervised settings.

\subsubsection{Zero-Shot Retrieval}
\label{ssec:rt-zero-shot}

Since edge questions are available to provide auxiliary information, edges can be directly integrated in the zero-shot retrieval process. Our motivation is straightforward: if there can be improvement with zero-shot retrieval, it ensures that these questions bring positive information gain, thus confirming the efficacy for augmenting local context.

Concretely, the hidden states (embeddings) for the query, nodes and edges are obtained by the encoder. Let $\mathbf{h}^v_i$ be the L2-normalized hidden state for the \textit{i}th node, $\mathbf{h}^e_{ij}$ for its \textit{j}th outgoing questions, $\mathbf{h}^q$ for the query. The interpolated similarity $\mathcal{S}_i$ between the query and $i$th candidate is defined as:
\begin{align}
\label{eq:rt}
    \mathcal{S} = \mathbf{h}^q \cdot \mathbf{h}^v_i + \lambda \cdot \max (\mathbf{h}^q \cdot \mathbf{h}^e_{ij}) |^M_{j=1}
\end{align}
The final similarity $\mathcal{S}$ is the typical query-node similarity interpolated with the query-edge similarity by $\lambda$, which is then the max query-question similarity out of total $M$ questions.  $\mathcal{S}$ among all nodes are then sorted for retrieval ranking, being a zero-shot approach without task-specific training.

\subsubsection{Supervised Learning}
\label{ssec:rt-supervised}

We then introduce our proposed supervised approach that reranks candidates with augmented node embeddings. Specifically, the enrichment is formulated as an attention, with the user query as \textit{query}, edge questions as both \textit{key} and \textit{value}, such that a new node embedding is obtained attending its edge questions conditioned on the query. Let $\mathcal{A}_i$ be the attention scores of the \textit{i}th candidate node, the augmented node embedding $\mathbf{h}^a_i$ is denoted as:
\begin{align}
    \mathcal{A}_i &= \text{softmax}\big(\frac{(\mathbf{h}^q W_Q)(\mathbf{h}^e_{ij} W_K)^T) }{\sqrt{d}}\big) \;|^M_{j=1}\\
    \mathbf{h}^a_i &= \mathbf{h}^v_i + \mathcal{A}_i \; (\mathbf{h}^e_{ij} W_V) |^M_{j=1}
\end{align}
$W_{Q/K/V}$ is the parameter for \textit{query}/\textit{key}/\textit{value} in attention, and $d$ is the \textit{query} dimension size. For a node $v_i$, we provide both outgoing and incoming questions to/from its direct neighbor node for bidirectional contextual information.

With the augmented embedding for the \textit{i}th node $\mathbf{h}^a_i$, the model simply reranks top retrieved candidates from a baseline system. It is trained with the supervised contrastive loss \cite{supervised-contrastive} to maximize the similarity between each query $q$ and its positive targets $P(q)$ among $N$ in-batch candidates (details in Appx~\ref{appx:exp}):
\begin{align}
    \mathcal{L} = \frac{-1}{|P(q)|} \sum_{x \in P(q)} \log \frac{\exp (\mathbf{h}^q \cdot \mathbf{h}^a_x)}{\sum^N_{y=1} \exp (\mathbf{h}^q \cdot \mathbf{h}^a_y)}
\end{align}

\subsection{Experiments}
\label{ssec:rt-exp}

\paragraph{Data}
For experiments situating our purpose, we adapt the data from \cite{xu2023plot} with slight modification. First, we use the available data of \textit{Notre-Dame de Paris} in Chinese for training and evaluation, instead of using all available novels to avoid large-scale graph realization. Second, the original task operates retrieval on sentence-level. Similar to Section~\ref{sec:task-recap}, we take short snippets as graph nodes, and label positive snippets converted from the original positive sentences.
The resulting dataset has 1288 candidate snippets in total, with 29484/1000/510 queries for the train/dev/test split.

\paragraph{Evaluation Metric}
A query may have one or many positive snippets (up to 7). We take the typical information retrieval metric normalized Discounted Cumulative Gain (nDCG), assigning the same relevance for each positive snippet equally.

\paragraph{Methods}
Four methods are evaluated as follows; all methods adopt BGE-Large encoder\footnote{\url{https://huggingface.co/BAAI/bge-large-zh-v1.5}}.
\begin{itemize}[noitemsep,nolistsep,leftmargin=*]
    \item Zero Shot (ZS): the zero-shot method that ranks candidates based on the query-node similarity.
    \item ZS+\graphns: our proposed interpolation with query-edge similarity; $\lambda$ is tuned on the dev set.
    \item Supervised (SU): the baseline supervised model without leveraging \graphns.
    \item SU+\graphns: our proposed rerank model that utilizes the global-contextualized embeddings; the inference reranks top 50 candidates by SU.
\end{itemize}

\begin{table}[htbp!]
\centering
\resizebox{0.74\columnwidth}{!}{
\begin{tabular}{l|ccc}
\toprule
& \multicolumn{3}{c}{nDCG} \\
\cmidrule{2-4}
& @1 & @5 & @10 \\
\midrule
Zero Shot & 17.06 & 20.83 & 23.97 \\
\; +\graphns & \bf 18.82 & \bf 23.83 & \bf 27.37 \\
\midrule
Supervised & 37.84 & 46.78 & 49.61 \\
\; +\graphns & \bf 40.20 & \bf 49.00 & \bf 51.33 \\
\bottomrule
\end{tabular}}
\caption{Evaluation results of zero-shot and supervised settings on our test set of the plot retrieval task. nDCG is evaluated on the top-1/5/10 retrieved candidates.}
\label{tab:rt-result}
\vspace{-2ex}
\end{table}

\subsection{Results}
\label{ssec:rt-result}

Table~\ref{tab:rt-result} shows the evaluation results of the four settings. Notably, our proposed zero-shot interpolation with query-edge similarity improves upon its baseline on all nDCG metrics, leading 3.4\% on nDCG@10 ($\lambda = 0.1$), confirming the positive information gain from edges that contribute useful contextual information. The same trend still holds up for the supervised model that learns enriched embeddings leveraging edge relations, especially by the 2.4\% improvement on nDCG@1. 

Overall, \graph is shown helpful towards the acquisition of better local representation, through the explicit relational dependencies beyond local context. The empirical results advocate the direction of fine-grained context modeling, which could foster a more nuanced comprehension.

\section{Study III: Broader Application}
\label{sec:task-qa}

Our last study sheds light on the potentials of graph utility in broader applications. As a first step towards this new direction, in this work, we evaluate with Retrieval-Augmented Generation (RAG) \cite{rag} in the task of long document question answering.
Experiments are conducted on QuALITY \cite{pang-etal-2022-quality}, a multi-choice QA dataset on narrative documents, mostly being fiction stories from Project Gutenberg. With an averaged length of 5k+ tokens per document, we adopt the retrieval-based approaches, where relevant snippets conditioned on the question are retrieved first, then fed to a LLM to generate answers, following a standard RAG paradigm.

Especially, QuALITY was constructed with global evidences in mind: questions may require multiple parts in the document to reason upon. 
Therefore, \graph may assist to recognize more relevant snippets through the extracted relations across the narrative context, leading to improved QA performance benefited from enhanced retrieval.

\paragraph{Methods}
Retrieval-based approaches are commonly adopted for tackling long context, which have been evaluated on QuALITY by previous works \cite{pang-etal-2022-quality,quality-baseline,sarthi2024raptor}.
Following these setup, we split the full document by short snippets, and retrieve relevant snippets with regard to the question, which are then concatenated as the shortened context for subsequent zero-shot QA inference by LLMs.
To leverage \graphns, we apply the same retrieval process described in Section~\ref{ssec:rt-zero-shot} to identify relevant snippets, where the query-edge similarity is interpolated as in Eq~\eqref{eq:rt} using BGE-Large encoder.

\paragraph{Experiments}
We employ Llama2 \cite{llama2} and ChatGPT for the zero-shot QA inference. As evaluation on the test set requires submission to the ZeroSCROLLS leaderboard \cite{zeroscrolls}, we first perform fine-grained performance analysis on the dev set with short retrieved context ($<$1k tokens), then submit the final test set results using ChatGPT with 1.5k context limit, aligned with \citet{quality-baseline} for direct comparison. The baseline retrieval method and our \textbf{E}nhanced retrieval are denoted by \textbf{\texttt{R}} and \textbf{\texttt{ER}} respectively.

Table~\ref{tab:qa-dev} \& \ref{tab:qa-test} present the evaluation results on the dev set and test set respectively.
Results on the dev set suggest that \texttt{ER} can boost QA performance with all LLMs, especially with the smaller 7B model by 5\% accuracy, fulfilling our initiative to effectively utilize \graph in broader applications. The improvement from enhanced context retrieval is consistent, further confirmed by the 2\% leading margin with ChatGPT on both the dev and test set.

Having demonstrated that \graph can improve RAG in narratives through enhanced retrieval, its utility beyond the retrieval process may be further exploited, e.g. potential facilitation on LLM pretraining or inference directly.
We leave future research to explore additional integration of fine-grained context modeling.

\begin{table}[tbp!]
\centering
\resizebox{\columnwidth}{!}{
\begin{tabular}{l|ccc}
\toprule
& \texttt{R} && \texttt{ER} \\
\midrule
Llama2-7B & 40.97 (± 0.67) & $\longrightarrow$ & 45.97 (± 0.63) \\
Llama2-70B & 61.56 (± 0.06) & $\longrightarrow$ & 63.98 (± 0.23) \\
ChatGPT & 63.66 (± 0.06) & $\longrightarrow$ & 65.92 (± 0.34) \\
\bottomrule
\end{tabular}}
\caption{Evaluation results on the dev set of QuALITY: accuracy with standard deviation (from three runs). Enhanced Retrieval (\texttt{ER}) improves QA consistently.}
\label{tab:qa-dev}
\end{table}

\begin{table}[tbp!]
\centering
\resizebox{0.95\columnwidth}{!}{
\begin{tabular}{lc|lc}
\toprule 
ChatGPT\textsuperscript{*} & 66.6 & ChatGPT (\texttt{R}) & 70.8 \\
Llama2-70B (\texttt{R})\textsuperscript{*} & 70.3 & ChatGPT (\texttt{ER}) & \bf 72.8 \\
\bottomrule
\end{tabular}}
\caption{Evaluation results on the test set of QuALITY submitted to the ZeroSCROLLS leaderboard. Accuracy of ChatGPT\textsuperscript{*} is provided by the ZeroSCROLLS organizers; Llama2-70B (\texttt{R})\textsuperscript{*} is reported by \citet{quality-baseline}.  Performance of three retrieval-based experiments are directly comparable (same 1.5k context limit). We exclude another related work RAPTOR \cite{sarthi2024raptor}, as they use smaller QA models and different context limit, thus not directly comparable.}
\label{tab:qa-test}
\vspace{-1ex}
\end{table}

\section{Conclusion}
\label{sec:conclusion}

We address the distinctive characteristics of narratives, and propose a novel paradigm of fine-grained context modeling, which explicitly captures the inter-connective coherence within narrative context.
A graph is thereby formulated, dubbed \graphns, with edges encompassing free-form retrospective questions to depict the relational dependencies.
\graph is practically realized by LLMs via our designed two-stage prompting scheme, leveraging the promising development of LLMs without reliance on human annotations.
To examine the graph properties and its utility, three unique studies are conducted, where \graph is shown to bring empirical improvement on various narrative applications.

\section*{Limitations}

While we have demonstrated the usefulness of our proposed \graphns, upon manually verifying the generated edge questions, deficiencies do exist in the current graph generation approach:
\begin{itemize}[noitemsep,nolistsep,leftmargin=*]
\item The generated questions are not free from noises, as mentioned in Section~\ref{sec:graph}. One common scenario occurs when pairs of context chunks are irrelevant to each other. GPT-4 struggles to accurately identify irrelevancy, leading it to ask questions that lack informativeness.
\item Our approach does not handle the scenario where there is joint dependency among three or more chunks. As we generate questions upon pairs, sometimes the key connecting information exists in the third chunk and is missing, preventing the recognition and formulation of useful questions.
\end{itemize}
Despite the aforementioned issues, our graph still proves beneficial in various applications. This is partly due to the fact that Large Language Models (LLMs) and our learned models possess the capability to automatically discern which information to utilize. Still, enhancing the quality of questions could further augment the benefits derived from our graph, highlighting the potentials of our proposed representation of narrative context.

An additional limitation lies in our filtering algorithm. For LLMs that struggle with following instructions accurately, the current filtering strategy may prove inadequate. For instance, if an LLM repeatedly poses questions that could be understood and answered solely by referring to prior texts, our filtering process is inefficiency to rule out these questions. One potential solution to mitigate this issue could involve implementing a matching model between the questions and the target texts. However, since our work employs GPT-4 alongside Chain-of-Thought, which effectively reduces such instances of shortcut-taking, we have opted to retain the current strategy. We acknowledge the possibility of exploring alternative LLMs with more sophisticated filtering strategies in future work.

\bibliography{acl}

\begin{thebibliography}{69}
\expandafter\ifx\csname natexlab\endcsname\relax\def\natexlab#1{#1}\fi

\bibitem[{Anagnostidis et~al.(2023)Anagnostidis, Pavllo, Biggio, Noci, Lucchi, and Hofmann}]{NEURIPS2023_cdaac2a0}
Sotiris Anagnostidis, Dario Pavllo, Luca Biggio, Lorenzo Noci, Aurelien Lucchi, and Thomas Hofmann. 2023.
\newblock \href {https://proceedings.neurips.cc/paper_files/paper/2023/file/cdaac2a02c4fdcae77ba083b110efcc3-Paper-Conference.pdf} {Dynamic context pruning for efficient and interpretable autoregressive transformers}.
\newblock In \emph{Advances in Neural Information Processing Systems}, volume~36, pages 65202--65223. Curran Associates, Inc.

\bibitem[{Benz and Jasinskaja(2017)}]{qud-survey}
Anton Benz and Katja Jasinskaja. 2017.
\newblock \href {https://doi.org/10.1080/0163853X.2017.1316038} {Questions under discussion: From sentence to discourse}.
\newblock \emph{Discourse Processes}, 54:177--186.

\bibitem[{Chen et~al.(2023{\natexlab{a}})Chen, Pasunuru, Weston, and Celikyilmaz}]{mem-walk}
Howard Chen, Ramakanth Pasunuru, Jason Weston, and Asli Celikyilmaz. 2023{\natexlab{a}}.
\newblock \href {http://arxiv.org/abs/2310.05029} {Walking down the memory maze: Beyond context limit through interactive reading}.

\bibitem[{Chen et~al.(2024)Chen, Han, and Zhang}]{chen2024comm}
Pei Chen, Boran Han, and Shuai Zhang. 2024.
\newblock \href {https://arxiv.org/abs/2404.17729} {Comm: Collaborative multi-agent, multi-reasoning-path prompting for complex problem solving}.
\newblock In \emph{Proceedings of the 2024 Conference of the North American Chapter of the Association for Computational Linguistics: Human Language Technologies}, Mexico City, Mexico. Association for Computational Linguistics.

\bibitem[{Chen et~al.(2023{\natexlab{b}})Chen, Wong, Chen, and Tian}]{chen2023extending}
Shouyuan Chen, Sherman Wong, Liangjian Chen, and Yuandong Tian. 2023{\natexlab{b}}.
\newblock \href {http://arxiv.org/abs/2306.15595} {Extending context window of large language models via positional interpolation}.

\bibitem[{Chevalier et~al.(2023)Chevalier, Wettig, Ajith, and Chen}]{chevalier-etal-2023-adapting}
Alexis Chevalier, Alexander Wettig, Anirudh Ajith, and Danqi Chen. 2023.
\newblock \href {https://doi.org/10.18653/v1/2023.emnlp-main.232} {Adapting language models to compress contexts}.
\newblock In \emph{Proceedings of the 2023 Conference on Empirical Methods in Natural Language Processing}, pages 3829--3846, Singapore. Association for Computational Linguistics.

\bibitem[{Choi et~al.(2021)Choi, Palomaki, Lamm, Kwiatkowski, Das, and Collins}]{decontextualization-choi-2021}
Eunsol Choi, Jennimaria Palomaki, Matthew Lamm, Tom Kwiatkowski, Dipanjan Das, and Michael Collins. 2021.
\newblock \href {https://doi.org/10.1162/tacl_a_00377} {Decontextualization: Making sentences stand-alone}.
\newblock \emph{Transactions of the Association for Computational Linguistics}, 9:447--461.

\bibitem[{Dai et~al.(2019)Dai, Yang, Yang, Carbonell, Le, and Salakhutdinov}]{dai-etal-2019-transformer}
Zihang Dai, Zhilin Yang, Yiming Yang, Jaime Carbonell, Quoc Le, and Ruslan Salakhutdinov. 2019.
\newblock \href {https://doi.org/10.18653/v1/P19-1285} {Transformer-{XL}: Attentive language models beyond a fixed-length context}.
\newblock In \emph{Proceedings of the 57th Annual Meeting of the Association for Computational Linguistics}, pages 2978--2988, Florence, Italy. Association for Computational Linguistics.

\bibitem[{De~Kuthy et~al.(2020)De~Kuthy, Kannan, Santhi~Ponnusamy, and Meurers}]{de-kuthy-etal-2020-towards}
Kordula De~Kuthy, Madeeswaran Kannan, Haemanth Santhi~Ponnusamy, and Detmar Meurers. 2020.
\newblock \href {https://doi.org/10.18653/v1/2020.coling-main.509} {Towards automatically generating questions under discussion to link information and discourse structure}.
\newblock In \emph{Proceedings of the 28th International Conference on Computational Linguistics}, pages 5786--5798, Barcelona, Spain (Online). International Committee on Computational Linguistics.

\bibitem[{De~Kuthy et~al.(2018)De~Kuthy, Reiter, and Riester}]{de-kuthy-etal-2018-qud}
Kordula De~Kuthy, Nils Reiter, and Arndt Riester. 2018.
\newblock \href {https://aclanthology.org/L18-1304} {{QUD}-based annotation of discourse structure and information structure: Tool and evaluation}.
\newblock In \emph{Proceedings of the Eleventh International Conference on Language Resources and Evaluation ({LREC} 2018)}, Miyazaki, Japan. European Language Resources Association (ELRA).

\bibitem[{Devlin et~al.(2019)Devlin, Chang, Lee, and Toutanova}]{bert}
Jacob Devlin, Ming-Wei Chang, Kenton Lee, and Kristina Toutanova. 2019.
\newblock \href {https://doi.org/10.18653/v1/N19-1423} {{BERT}: Pre-training of deep bidirectional transformers for language understanding}.
\newblock In \emph{Proceedings of the 2019 Conference of the North {A}merican Chapter of the Association for Computational Linguistics: Human Language Technologies, Volume 1 (Long and Short Papers)}, pages 4171--4186, Minneapolis, Minnesota. Association for Computational Linguistics.

\bibitem[{Ding et~al.(2019)Ding, Zhou, Chen, Yang, and Tang}]{ding2019cognitive}
Ming Ding, Chang Zhou, Qibin Chen, Hongxia Yang, and Jie Tang. 2019.
\newblock Cognitive graph for multi-hop reading comprehension at scale.
\newblock \emph{arXiv preprint arXiv:1905.05460}.

\bibitem[{Du et~al.(2017)Du, Shao, and Cardie}]{du-etal-2017-learning}
Xinya Du, Junru Shao, and Claire Cardie. 2017.
\newblock \href {https://doi.org/10.18653/v1/P17-1123} {Learning to ask: Neural question generation for reading comprehension}.
\newblock In \emph{Proceedings of the 55th Annual Meeting of the Association for Computational Linguistics (Volume 1: Long Papers)}, pages 1342--1352, Vancouver, Canada. Association for Computational Linguistics.

\bibitem[{Ge et~al.(2024{\natexlab{a}})Ge, Zhang, Liu, Zhang, Han, and Gao}]{ge2024model}
Suyu Ge, Yunan Zhang, Liyuan Liu, Minjia Zhang, Jiawei Han, and Jianfeng Gao. 2024{\natexlab{a}}.
\newblock \href {https://openreview.net/forum?id=uNrFpDPMyo} {Model tells you what to discard: Adaptive {KV} cache compression for {LLM}s}.
\newblock In \emph{The Twelfth International Conference on Learning Representations}.

\bibitem[{Ge et~al.(2024{\natexlab{b}})Ge, Jing, Wang, Wang, Chen, and Wei}]{ge2024incontext}
Tao Ge, Hu~Jing, Lei Wang, Xun Wang, Si-Qing Chen, and Furu Wei. 2024{\natexlab{b}}.
\newblock \href {https://openreview.net/forum?id=uREj4ZuGJE} {In-context autoencoder for context compression in a large language model}.
\newblock In \emph{The Twelfth International Conference on Learning Representations}.

\bibitem[{Graesser et~al.(1994)Graesser, Singer, and Trabasso}]{backward2}
Arthur Graesser, Murray Singer, and Tom Trabasso. 1994.
\newblock \href {https://doi.org/10.1037/0033-295X.101.3.371} {Constructing inferences during narrative text comprehension}.
\newblock \emph{Psychological review}, 101:371--95.

\bibitem[{Grosz and Sidner(1986)}]{discourse}
Barbara~J. Grosz and Candace~L. Sidner. 1986.
\newblock \href {https://aclanthology.org/J86-3001} {Attention, intentions, and the structure of discourse}.
\newblock \emph{Computational Linguistics}, 12(3):175--204.

\bibitem[{Hu and Wan(2023)}]{hu-wan-2023-exploring}
Xinyu Hu and Xiaojun Wan. 2023.
\newblock \href {https://doi.org/10.18653/v1/2023.emnlp-main.857} {Exploring discourse structure in document-level machine translation}.
\newblock In \emph{Proceedings of the 2023 Conference on Empirical Methods in Natural Language Processing}, pages 13889--13902, Singapore. Association for Computational Linguistics.

\bibitem[{Ji and Smith(2017)}]{ji-smith-2017-neural}
Yangfeng Ji and Noah~A. Smith. 2017.
\newblock \href {https://doi.org/10.18653/v1/P17-1092} {Neural discourse structure for text categorization}.
\newblock In \emph{Proceedings of the 55th Annual Meeting of the Association for Computational Linguistics (Volume 1: Long Papers)}, pages 996--1005, Vancouver, Canada. Association for Computational Linguistics.

\bibitem[{Jiang et~al.(2024)Jiang, Sablayrolles, Roux, Mensch, Savary, Bamford, Chaplot, de~las Casas, Hanna, Bressand, Lengyel, Bour, Lample, Lavaud, Saulnier, Lachaux, Stock, Subramanian, Yang, Antoniak, Scao, Gervet, Lavril, Wang, Lacroix, and Sayed}]{jiang2024mixtral}
Albert~Q. Jiang, Alexandre Sablayrolles, Antoine Roux, Arthur Mensch, Blanche Savary, Chris Bamford, Devendra~Singh Chaplot, Diego de~las Casas, Emma~Bou Hanna, Florian Bressand, Gianna Lengyel, Guillaume Bour, Guillaume Lample, Lélio~Renard Lavaud, Lucile Saulnier, Marie-Anne Lachaux, Pierre Stock, Sandeep Subramanian, Sophia Yang, Szymon Antoniak, Teven~Le Scao, Théophile Gervet, Thibaut Lavril, Thomas Wang, Timothée Lacroix, and William~El Sayed. 2024.
\newblock \href {http://arxiv.org/abs/2401.04088} {Mixtral of experts}.

\bibitem[{Khosla et~al.(2020)Khosla, Teterwak, Wang, Sarna, Tian, Isola, Maschinot, Liu, and Krishnan}]{supervised-contrastive}
Prannay Khosla, Piotr Teterwak, Chen Wang, Aaron Sarna, Yonglong Tian, Phillip Isola, Aaron Maschinot, Ce~Liu, and Dilip Krishnan. 2020.
\newblock \href {https://proceedings.neurips.cc/paper/2020/file/d89a66c7c80a29b1bdbab0f2a1a94af8-Paper.pdf} {Supervised contrastive learning}.
\newblock In \emph{Advances in Neural Information Processing Systems}, volume~33, pages 18661--18673. Curran Associates, Inc.

\bibitem[{Ko et~al.(2020)Ko, Chen, Huang, Durrett, and Li}]{ko-etal-2020-inquisitive}
Wei-Jen Ko, Te-yuan Chen, Yiyan Huang, Greg Durrett, and Junyi~Jessy Li. 2020.
\newblock \href {https://doi.org/10.18653/v1/2020.emnlp-main.530} {Inquisitive question generation for high level text comprehension}.
\newblock In \emph{Proceedings of the 2020 Conference on Empirical Methods in Natural Language Processing (EMNLP)}, pages 6544--6555, Online. Association for Computational Linguistics.

\bibitem[{Ko et~al.(2022)Ko, Dalton, Simmons, Fisher, Durrett, and Li}]{ko-etal-2022-discourse}
Wei-Jen Ko, Cutter Dalton, Mark Simmons, Eliza Fisher, Greg Durrett, and Junyi~Jessy Li. 2022.
\newblock \href {https://doi.org/10.18653/v1/2022.emnlp-main.806} {Discourse comprehension: A question answering framework to represent sentence connections}.
\newblock In \emph{Proceedings of the 2022 Conference on Empirical Methods in Natural Language Processing}, pages 11752--11764, Abu Dhabi, United Arab Emirates. Association for Computational Linguistics.

\bibitem[{Ko et~al.(2023)Ko, Wu, Dalton, Srinivas, Durrett, and Li}]{ko-etal-2023-discourse}
Wei-Jen Ko, Yating Wu, Cutter Dalton, Dananjay Srinivas, Greg Durrett, and Junyi~Jessy Li. 2023.
\newblock \href {https://doi.org/10.18653/v1/2023.findings-acl.710} {Discourse analysis via questions and answers: Parsing dependency structures of questions under discussion}.
\newblock In \emph{Findings of the Association for Computational Linguistics: ACL 2023}, pages 11181--11195, Toronto, Canada. Association for Computational Linguistics.

\bibitem[{Ko{\v{c}}isk{\'y} et~al.(2018)Ko{\v{c}}isk{\'y}, Schwarz, Blunsom, Dyer, Hermann, Melis, and Grefenstette}]{kocisky-etal-2018-narrativeqa}
Tom{\'a}{\v{s}} Ko{\v{c}}isk{\'y}, Jonathan Schwarz, Phil Blunsom, Chris Dyer, Karl~Moritz Hermann, G{\'a}bor Melis, and Edward Grefenstette. 2018.
\newblock \href {https://doi.org/10.1162/tacl_a_00023} {The {N}arrative{QA} reading comprehension challenge}.
\newblock \emph{Transactions of the Association for Computational Linguistics}, 6:317--328.

\bibitem[{Kuppevelt(1995)}]{qud_1995}
Jan~Van Kuppevelt. 1995.
\newblock \href {https://doi.org/10.1017/S002222670000058X} {Discourse structure, topicality and questioning}.
\newblock \emph{Journal of Linguistics}, 31(1):109–147.

\bibitem[{Lal et~al.(2021)Lal, Chambers, Mooney, and Balasubramanian}]{lal-etal-2021-tellmewhy}
Yash~Kumar Lal, Nathanael Chambers, Raymond Mooney, and Niranjan Balasubramanian. 2021.
\newblock \href {https://doi.org/10.18653/v1/2021.findings-acl.53} {{T}ell{M}e{W}hy: A dataset for answering why-questions in narratives}.
\newblock In \emph{Findings of the Association for Computational Linguistics: ACL-IJCNLP 2021}, pages 596--610, Online. Association for Computational Linguistics.

\bibitem[{Lewis et~al.(2020)Lewis, Perez, Piktus, Petroni, Karpukhin, Goyal, K\"{u}ttler, Lewis, Yih, Rockt\"{a}schel, Riedel, and Kiela}]{rag}
Patrick Lewis, Ethan Perez, Aleksandra Piktus, Fabio Petroni, Vladimir Karpukhin, Naman Goyal, Heinrich K\"{u}ttler, Mike Lewis, Wen-tau Yih, Tim Rockt\"{a}schel, Sebastian Riedel, and Douwe Kiela. 2020.
\newblock \href {https://proceedings.neurips.cc/paper_files/paper/2020/file/6b493230205f780e1bc26945df7481e5-Paper.pdf} {Retrieval-augmented generation for knowledge-intensive nlp tasks}.
\newblock In \emph{Advances in Neural Information Processing Systems}, volume~33, pages 9459--9474. Curran Associates, Inc.

\bibitem[{Li et~al.(2024)Li, Wang, Xu, Pang, Yu, Lin, Wang, and Zhou}]{recap}
Jiangnan Li, Qiujing Wang, Liyan Xu, Wenjie Pang, Mo~Yu, Zheng Lin, Weiping Wang, and Jie Zhou. 2024.
\newblock \href {http://arxiv.org/abs/2402.07271} {Previously on the stories: Recap snippet identification for story reading}.

\bibitem[{Li et~al.(2021)Li, Ma, Yu, Wu, Gao, Ji, and McKeown}]{li2021timeline}
Manling Li, Tengfei Ma, Mo~Yu, Lingfei Wu, Tian Gao, Heng Ji, and Kathleen McKeown. 2021.
\newblock Timeline summarization based on event graph compression via time-aware optimal transport.
\newblock In \emph{Proceedings of EMNLP 2021}, pages 6443--6456.

\bibitem[{Li et~al.(2020)Li, Zeng, Lin, Cho, Ji, May, Chambers, and Voss}]{li2020connecting}
Manling Li, Qi~Zeng, Ying Lin, Kyunghyun Cho, Heng Ji, Jonathan May, Nathanael Chambers, and Clare Voss. 2020.
\newblock Connecting the dots: Event graph schema induction with path language modeling.
\newblock In \emph{Proceedings of EMNLP 2020}, pages 684--695.

\bibitem[{Mann and Thompson(1988)}]{rst}
William Mann and Sandra Thompson. 1988.
\newblock \href {https://doi.org/10.1515/text.1.1988.8.3.243} {Rethorical structure theory: Toward a functional theory of text organization}.
\newblock \emph{Text}, 8:243--281.

\bibitem[{Munkhdalai et~al.(2024)Munkhdalai, Faruqui, and Gopal}]{munkhdalai2024leave}
Tsendsuren Munkhdalai, Manaal Faruqui, and Siddharth Gopal. 2024.
\newblock \href {http://arxiv.org/abs/2404.07143} {Leave no context behind: Efficient infinite context transformers with infini-attention}.

\bibitem[{Nair et~al.(2023)Nair, Somasundaram, Saxena, and Goswami}]{nair-etal-2023-drilling}
Inderjeet Nair, Shwetha Somasundaram, Apoorv Saxena, and Koustava Goswami. 2023.
\newblock \href {https://doi.org/10.18653/v1/2023.findings-emnlp.972} {Drilling down into the discourse structure with {LLM}s for long document question answering}.
\newblock In \emph{Findings of the Association for Computational Linguistics: EMNLP 2023}, pages 14593--14606, Singapore. Association for Computational Linguistics.

\bibitem[{Newman et~al.(2023)Newman, Soldaini, Fok, Cohan, and Lo}]{decontextualization-newman-2023}
Benjamin Newman, Luca Soldaini, Raymond Fok, Arman Cohan, and Kyle Lo. 2023.
\newblock \href {https://doi.org/10.18653/v1/2023.emnlp-main.193} {A question answering framework for decontextualizing user-facing snippets from scientific documents}.
\newblock In \emph{Proceedings of the 2023 Conference on Empirical Methods in Natural Language Processing}, pages 3194--3212, Singapore. Association for Computational Linguistics.

\bibitem[{Nguyen et~al.(2022)Nguyen, Min, Dernoncourt, and Nguyen}]{nguyen-etal-2022-joint}
Minh~Van Nguyen, Bonan Min, Franck Dernoncourt, and Thien Nguyen. 2022.
\newblock \href {https://doi.org/10.18653/v1/2022.naacl-main.324} {Joint extraction of entities, relations, and events via modeling inter-instance and inter-label dependencies}.
\newblock In \emph{Proceedings of the 2022 Conference of the North American Chapter of the Association for Computational Linguistics: Human Language Technologies}, pages 4363--4374, Seattle, United States. Association for Computational Linguistics.

\bibitem[{OpenAI et~al.(2024)OpenAI, Achiam, Adler, Agarwal, Ahmad, Akkaya, Aleman, Almeida, Altenschmidt, Altman, Anadkat, Avila, Babuschkin, Balaji, Balcom, Baltescu, Bao, Bavarian, Belgum, Bello, Berdine, Bernadett-Shapiro, Berner, Bogdonoff, Boiko, Boyd, Brakman, Brockman, Brooks, Brundage, Button, Cai, Campbell, Cann, Carey, Carlson, Carmichael, Chan, Chang, Chantzis, Chen, Chen, Chen, Chen, Chen, Chess, Cho, Chu, Chung, Cummings, Currier, Dai, Decareaux, Degry, Deutsch, Deville, Dhar, Dohan, Dowling, Dunning, Ecoffet, Eleti, Eloundou, Farhi, Fedus, Felix, Fishman, Forte, Fulford, Gao, Georges, Gibson, Goel, Gogineni, Goh, Gontijo-Lopes, Gordon, Grafstein, Gray, Greene, Gross, Gu, Guo, Hallacy, Han, Harris, He, Heaton, Heidecke, Hesse, Hickey, Hickey, Hoeschele, Houghton, Hsu, Hu, Hu, Huizinga, Jain, Jain, Jang, Jiang, Jiang, Jin, Jin, Jomoto, Jonn, Jun, Kaftan, Łukasz Kaiser, Kamali, Kanitscheider, Keskar, Khan, Kilpatrick, Kim, Kim, Kim, Kirchner, Kiros, Knight, Kokotajlo, Łukasz Kondraciuk,
  Kondrich, Konstantinidis, Kosic, Krueger, Kuo, Lampe, Lan, Lee, Leike, Leung, Levy, Li, Lim, Lin, Lin, Litwin, Lopez, Lowe, Lue, Makanju, Malfacini, Manning, Markov, Markovski, Martin, Mayer, Mayne, McGrew, McKinney, McLeavey, McMillan, McNeil, Medina, Mehta, Menick, Metz, Mishchenko, Mishkin, Monaco, Morikawa, Mossing, Mu, Murati, Murk, Mély, Nair, Nakano, Nayak, Neelakantan, Ngo, Noh, Ouyang, O'Keefe, Pachocki, Paino, Palermo, Pantuliano, Parascandolo, Parish, Parparita, Passos, Pavlov, Peng, Perelman, de~Avila Belbute~Peres, Petrov, de~Oliveira~Pinto, Michael, Pokorny, Pokrass, Pong, Powell, Power, Power, Proehl, Puri, Radford, Rae, Ramesh, Raymond, Real, Rimbach, Ross, Rotsted, Roussez, Ryder, Saltarelli, Sanders, Santurkar, Sastry, Schmidt, Schnurr, Schulman, Selsam, Sheppard, Sherbakov, Shieh, Shoker, Shyam, Sidor, Sigler, Simens, Sitkin, Slama, Sohl, Sokolowsky, Song, Staudacher, Such, Summers, Sutskever, Tang, Tezak, Thompson, Tillet, Tootoonchian, Tseng, Tuggle, Turley, Tworek, Uribe, Vallone,
  Vijayvergiya, Voss, Wainwright, Wang, Wang, Wang, Ward, Wei, Weinmann, Welihinda, Welinder, Weng, Weng, Wiethoff, Willner, Winter, Wolrich, Wong, Workman, Wu, Wu, Wu, Xiao, Xu, Yoo, Yu, Yuan, Zaremba, Zellers, Zhang, Zhang, Zhao, Zheng, Zhuang, Zhuk, and Zoph}]{openai2024gpt4}
OpenAI, Josh Achiam, Steven Adler, Sandhini Agarwal, Lama Ahmad, Ilge Akkaya, Florencia~Leoni Aleman, Diogo Almeida, Janko Altenschmidt, Sam Altman, Shyamal Anadkat, Red Avila, Igor Babuschkin, Suchir Balaji, Valerie Balcom, Paul Baltescu, Haiming Bao, Mohammad Bavarian, Jeff Belgum, Irwan Bello, Jake Berdine, Gabriel Bernadett-Shapiro, Christopher Berner, Lenny Bogdonoff, Oleg Boiko, Madelaine Boyd, Anna-Luisa Brakman, Greg Brockman, Tim Brooks, Miles Brundage, Kevin Button, Trevor Cai, Rosie Campbell, Andrew Cann, Brittany Carey, Chelsea Carlson, Rory Carmichael, Brooke Chan, Che Chang, Fotis Chantzis, Derek Chen, Sully Chen, Ruby Chen, Jason Chen, Mark Chen, Ben Chess, Chester Cho, Casey Chu, Hyung~Won Chung, Dave Cummings, Jeremiah Currier, Yunxing Dai, Cory Decareaux, Thomas Degry, Noah Deutsch, Damien Deville, Arka Dhar, David Dohan, Steve Dowling, Sheila Dunning, Adrien Ecoffet, Atty Eleti, Tyna Eloundou, David Farhi, Liam Fedus, Niko Felix, Simón~Posada Fishman, Juston Forte, Isabella Fulford, Leo
  Gao, Elie Georges, Christian Gibson, Vik Goel, Tarun Gogineni, Gabriel Goh, Rapha Gontijo-Lopes, Jonathan Gordon, Morgan Grafstein, Scott Gray, Ryan Greene, Joshua Gross, Shixiang~Shane Gu, Yufei Guo, Chris Hallacy, Jesse Han, Jeff Harris, Yuchen He, Mike Heaton, Johannes Heidecke, Chris Hesse, Alan Hickey, Wade Hickey, Peter Hoeschele, Brandon Houghton, Kenny Hsu, Shengli Hu, Xin Hu, Joost Huizinga, Shantanu Jain, Shawn Jain, Joanne Jang, Angela Jiang, Roger Jiang, Haozhun Jin, Denny Jin, Shino Jomoto, Billie Jonn, Heewoo Jun, Tomer Kaftan, Łukasz Kaiser, Ali Kamali, Ingmar Kanitscheider, Nitish~Shirish Keskar, Tabarak Khan, Logan Kilpatrick, Jong~Wook Kim, Christina Kim, Yongjik Kim, Jan~Hendrik Kirchner, Jamie Kiros, Matt Knight, Daniel Kokotajlo, Łukasz Kondraciuk, Andrew Kondrich, Aris Konstantinidis, Kyle Kosic, Gretchen Krueger, Vishal Kuo, Michael Lampe, Ikai Lan, Teddy Lee, Jan Leike, Jade Leung, Daniel Levy, Chak~Ming Li, Rachel Lim, Molly Lin, Stephanie Lin, Mateusz Litwin, Theresa Lopez, Ryan
  Lowe, Patricia Lue, Anna Makanju, Kim Malfacini, Sam Manning, Todor Markov, Yaniv Markovski, Bianca Martin, Katie Mayer, Andrew Mayne, Bob McGrew, Scott~Mayer McKinney, Christine McLeavey, Paul McMillan, Jake McNeil, David Medina, Aalok Mehta, Jacob Menick, Luke Metz, Andrey Mishchenko, Pamela Mishkin, Vinnie Monaco, Evan Morikawa, Daniel Mossing, Tong Mu, Mira Murati, Oleg Murk, David Mély, Ashvin Nair, Reiichiro Nakano, Rajeev Nayak, Arvind Neelakantan, Richard Ngo, Hyeonwoo Noh, Long Ouyang, Cullen O'Keefe, Jakub Pachocki, Alex Paino, Joe Palermo, Ashley Pantuliano, Giambattista Parascandolo, Joel Parish, Emy Parparita, Alex Passos, Mikhail Pavlov, Andrew Peng, Adam Perelman, Filipe de~Avila Belbute~Peres, Michael Petrov, Henrique~Ponde de~Oliveira~Pinto, Michael, Pokorny, Michelle Pokrass, Vitchyr~H. Pong, Tolly Powell, Alethea Power, Boris Power, Elizabeth Proehl, Raul Puri, Alec Radford, Jack Rae, Aditya Ramesh, Cameron Raymond, Francis Real, Kendra Rimbach, Carl Ross, Bob Rotsted, Henri Roussez,
  Nick Ryder, Mario Saltarelli, Ted Sanders, Shibani Santurkar, Girish Sastry, Heather Schmidt, David Schnurr, John Schulman, Daniel Selsam, Kyla Sheppard, Toki Sherbakov, Jessica Shieh, Sarah Shoker, Pranav Shyam, Szymon Sidor, Eric Sigler, Maddie Simens, Jordan Sitkin, Katarina Slama, Ian Sohl, Benjamin Sokolowsky, Yang Song, Natalie Staudacher, Felipe~Petroski Such, Natalie Summers, Ilya Sutskever, Jie Tang, Nikolas Tezak, Madeleine~B. Thompson, Phil Tillet, Amin Tootoonchian, Elizabeth Tseng, Preston Tuggle, Nick Turley, Jerry Tworek, Juan Felipe~Cerón Uribe, Andrea Vallone, Arun Vijayvergiya, Chelsea Voss, Carroll Wainwright, Justin~Jay Wang, Alvin Wang, Ben Wang, Jonathan Ward, Jason Wei, CJ~Weinmann, Akila Welihinda, Peter Welinder, Jiayi Weng, Lilian Weng, Matt Wiethoff, Dave Willner, Clemens Winter, Samuel Wolrich, Hannah Wong, Lauren Workman, Sherwin Wu, Jeff Wu, Michael Wu, Kai Xiao, Tao Xu, Sarah Yoo, Kevin Yu, Qiming Yuan, Wojciech Zaremba, Rowan Zellers, Chong Zhang, Marvin Zhang, Shengjia
  Zhao, Tianhao Zheng, Juntang Zhuang, William Zhuk, and Barret Zoph. 2024.
\newblock \href {http://arxiv.org/abs/2303.08774} {Gpt-4 technical report}.

\bibitem[{Pang et~al.(2022)Pang, Parrish, Joshi, Nangia, Phang, Chen, Padmakumar, Ma, Thompson, He, and Bowman}]{pang-etal-2022-quality}
Richard~Yuanzhe Pang, Alicia Parrish, Nitish Joshi, Nikita Nangia, Jason Phang, Angelica Chen, Vishakh Padmakumar, Johnny Ma, Jana Thompson, He~He, and Samuel Bowman. 2022.
\newblock \href {https://doi.org/10.18653/v1/2022.naacl-main.391} {{Q}u{ALITY}: Question answering with long input texts, yes!}
\newblock In \emph{Proceedings of the 2022 Conference of the North American Chapter of the Association for Computational Linguistics: Human Language Technologies}, pages 5336--5358, Seattle, United States. Association for Computational Linguistics.

\bibitem[{Peng et~al.(2024)Peng, Quesnelle, Fan, and Shippole}]{yarn}
Bowen Peng, Jeffrey Quesnelle, Honglu Fan, and Enrico Shippole. 2024.
\newblock \href {https://openreview.net/forum?id=wHBfxhZu1u} {Ya{RN}: Efficient context window extension of large language models}.
\newblock In \emph{The Twelfth International Conference on Learning Representations}.

\bibitem[{Prasad et~al.(2008)Prasad, Dinesh, Lee, Miltsakaki, Robaldo, Joshi, and Webber}]{pdtb}
Rashmi Prasad, Nikhil Dinesh, Alan Lee, Eleni Miltsakaki, Livio Robaldo, Aravind Joshi, and Bonnie Webber. 2008.
\newblock \href {http://www.lrec-conf.org/proceedings/lrec2008/pdf/754_paper.pdf} {The {P}enn {D}iscourse {T}ree{B}ank 2.0.}
\newblock In \emph{Proceedings of the Sixth International Conference on Language Resources and Evaluation ({LREC}'08)}, Marrakech, Morocco. European Language Resources Association (ELRA).

\bibitem[{Reimers and Gurevych(2019)}]{reimers-gurevych-2019-sentence}
Nils Reimers and Iryna Gurevych. 2019.
\newblock \href {https://doi.org/10.18653/v1/D19-1410} {Sentence-{BERT}: Sentence embeddings using {S}iamese {BERT}-networks}.
\newblock In \emph{Proceedings of the 2019 Conference on Empirical Methods in Natural Language Processing and the 9th International Joint Conference on Natural Language Processing (EMNLP-IJCNLP)}, pages 3982--3992, Hong Kong, China. Association for Computational Linguistics.

\bibitem[{Roberts(1996)}]{qud_1996}
Craige Roberts. 1996.
\newblock \href {https://doi.org/10.3765/sp.5.6} {Information structure in discourse: Towards an integrated formal theory of pragmatics}.
\newblock \emph{Journal of Heuristics - HEURISTICS}, 49.

\bibitem[{Sang et~al.(2022)Sang, Mou, Yu, Yao, Li, and Stanton}]{sang-etal-2022-tvshowguess}
Yisi Sang, Xiangyang Mou, Mo~Yu, Shunyu Yao, Jing Li, and Jeffrey Stanton. 2022.
\newblock \href {https://doi.org/10.18653/v1/2022.naacl-main.317} {{TVS}how{G}uess: Character comprehension in stories as speaker guessing}.
\newblock In \emph{Proceedings of the 2022 Conference of the North American Chapter of the Association for Computational Linguistics: Human Language Technologies}, pages 4267--4287, Seattle, United States. Association for Computational Linguistics.

\bibitem[{Sarthi et~al.(2024)Sarthi, Abdullah, Tuli, Khanna, Goldie, and Manning}]{sarthi2024raptor}
Parth Sarthi, Salman Abdullah, Aditi Tuli, Shubh Khanna, Anna Goldie, and Christopher~D Manning. 2024.
\newblock \href {https://openreview.net/forum?id=GN921JHCRw} {{RAPTOR}: Recursive abstractive processing for tree-organized retrieval}.
\newblock In \emph{The Twelfth International Conference on Learning Representations}.

\bibitem[{Shaham et~al.(2023)Shaham, Ivgi, Efrat, Berant, and Levy}]{zeroscrolls}
Uri Shaham, Maor Ivgi, Avia Efrat, Jonathan Berant, and Omer Levy. 2023.
\newblock \href {https://doi.org/10.18653/v1/2023.findings-emnlp.536} {{Z}ero{SCROLLS}: A zero-shot benchmark for long text understanding}.
\newblock In \emph{Findings of the Association for Computational Linguistics: EMNLP 2023}, pages 7977--7989, Singapore. Association for Computational Linguistics.

\bibitem[{Song et~al.(2020)Song, Park, Park, and Shim}]{backward3}
Hayoung Song, Bo-Yong Park, Hyunjin Park, and Won Shim. 2020.
\newblock \href {https://doi.org/10.1101/2020.07.10.194647} {Cognitive and neural state dynamics of story comprehension}.
\newblock \emph{Journal of Neuroscience}.

\bibitem[{Strubell et~al.(2018)Strubell, Verga, Andor, Weiss, and McCallum}]{strubell-etal-2018-linguistically}
Emma Strubell, Patrick Verga, Daniel Andor, David Weiss, and Andrew McCallum. 2018.
\newblock \href {https://doi.org/10.18653/v1/D18-1548} {Linguistically-informed self-attention for semantic role labeling}.
\newblock In \emph{Proceedings of the 2018 Conference on Empirical Methods in Natural Language Processing}, pages 5027--5038, Brussels, Belgium. Association for Computational Linguistics.

\bibitem[{Su et~al.(2024)Su, Xu, Xu, Li, and Huangfu}]{sig}
Zhenlin Su, Liyan Xu, Jin Xu, Jiangnan Li, and Mingdu Huangfu. 2024.
\newblock Sig: Speaker identification in literature via prompt-based generation.
\newblock \emph{Proceedings of the AAAI Conference on Artificial Intelligence}.

\bibitem[{Sun et~al.(2024)Sun, Xu, Yu, Chen, He, Zhao, and Liu}]{sun2024itd}
Wangtao Sun, Haotian Xu, Xuanqing Yu, Pei Chen, Shizhu He, Jun Zhao, and Kang Liu. 2024.
\newblock \href {https://arxiv.org/abs/2403.05789} {Itd: Large language models can teach themselves induction through deduction}.
\newblock In \emph{Proceedings of the 62nd Annual Meeting of the Association for Computational Linguistics}, Bangkok, Thailand. Association for Computational Linguistics.

\bibitem[{Thai et~al.(2022)Thai, Chang, Krishna, and Iyyer}]{thai-etal-2022-relic}
Katherine Thai, Yapei Chang, Kalpesh Krishna, and Mohit Iyyer. 2022.
\newblock \href {https://doi.org/10.18653/v1/2022.acl-long.517} {{REL}i{C}: Retrieving evidence for literary claims}.
\newblock In \emph{Proceedings of the 60th Annual Meeting of the Association for Computational Linguistics (Volume 1: Long Papers)}, pages 7500--7518, Dublin, Ireland. Association for Computational Linguistics.

\bibitem[{Touvron et~al.(2023)Touvron, Martin, Stone, Albert, Almahairi, Babaei, Bashlykov, Batra, Bhargava, Bhosale, Bikel, Blecher, Ferrer, Chen, Cucurull, Esiobu, Fernandes, Fu, Fu, Fuller, Gao, Goswami, Goyal, Hartshorn, Hosseini, Hou, Inan, Kardas, Kerkez, Khabsa, Kloumann, Korenev, Koura, Lachaux, Lavril, Lee, Liskovich, Lu, Mao, Martinet, Mihaylov, Mishra, Molybog, Nie, Poulton, Reizenstein, Rungta, Saladi, Schelten, Silva, Smith, Subramanian, Tan, Tang, Taylor, Williams, Kuan, Xu, Yan, Zarov, Zhang, Fan, Kambadur, Narang, Rodriguez, Stojnic, Edunov, and Scialom}]{llama2}
Hugo Touvron, Louis Martin, Kevin Stone, Peter Albert, Amjad Almahairi, Yasmine Babaei, Nikolay Bashlykov, Soumya Batra, Prajjwal Bhargava, Shruti Bhosale, Dan Bikel, Lukas Blecher, Cristian~Canton Ferrer, Moya Chen, Guillem Cucurull, David Esiobu, Jude Fernandes, Jeremy Fu, Wenyin Fu, Brian Fuller, Cynthia Gao, Vedanuj Goswami, Naman Goyal, Anthony Hartshorn, Saghar Hosseini, Rui Hou, Hakan Inan, Marcin Kardas, Viktor Kerkez, Madian Khabsa, Isabel Kloumann, Artem Korenev, Punit~Singh Koura, Marie-Anne Lachaux, Thibaut Lavril, Jenya Lee, Diana Liskovich, Yinghai Lu, Yuning Mao, Xavier Martinet, Todor Mihaylov, Pushkar Mishra, Igor Molybog, Yixin Nie, Andrew Poulton, Jeremy Reizenstein, Rashi Rungta, Kalyan Saladi, Alan Schelten, Ruan Silva, Eric~Michael Smith, Ranjan Subramanian, Xiaoqing~Ellen Tan, Binh Tang, Ross Taylor, Adina Williams, Jian~Xiang Kuan, Puxin Xu, Zheng Yan, Iliyan Zarov, Yuchen Zhang, Angela Fan, Melanie Kambadur, Sharan Narang, Aurelien Rodriguez, Robert Stojnic, Sergey Edunov, and Thomas
  Scialom. 2023.
\newblock \href {http://arxiv.org/abs/2307.09288} {Llama 2: Open foundation and fine-tuned chat models}.

\bibitem[{Trabasso and Sperry(1985)}]{backward1}
Tom Trabasso and Linda~L Sperry. 1985.
\newblock \href {https://doi.org/https://doi.org/10.1016/0749-596X(85)90048-8} {Causal relatedness and importance of story events}.
\newblock \emph{Journal of Memory and Language}, 24(5):595--611.

\bibitem[{Wang et~al.(2023)Wang, Dong, Cheng, Liu, Yan, Gao, and Wei}]{long-mem}
Weizhi Wang, Li~Dong, Hao Cheng, Xiaodong Liu, Xifeng Yan, Jianfeng Gao, and Furu Wei. 2023.
\newblock \href {https://openreview.net/forum?id=BryMFPQ4L6} {Augmenting language models with long-term memory}.
\newblock In \emph{Thirty-seventh Conference on Neural Information Processing Systems}.

\bibitem[{Wei et~al.(2022)Wei, Wang, Schuurmans, Bosma, ichter, Xia, Chi, Le, and Zhou}]{cot}
Jason Wei, Xuezhi Wang, Dale Schuurmans, Maarten Bosma, brian ichter, Fei Xia, Ed~Chi, Quoc~V Le, and Denny Zhou. 2022.
\newblock \href {https://proceedings.neurips.cc/paper_files/paper/2022/file/9d5609613524ecf4f15af0f7b31abca4-Paper-Conference.pdf} {Chain-of-thought prompting elicits reasoning in large language models}.
\newblock In \emph{Advances in Neural Information Processing Systems}, volume~35, pages 24824--24837. Curran Associates, Inc.

\bibitem[{Westera et~al.(2020)Westera, Mayol, and Rohde}]{westera-etal-2020-ted}
Matthijs Westera, Laia Mayol, and Hannah Rohde. 2020.
\newblock \href {https://aclanthology.org/2020.lrec-1.141} {{TED}-{Q}: {TED} talks and the questions they evoke}.
\newblock In \emph{Proceedings of the Twelfth Language Resources and Evaluation Conference}, pages 1118--1127, Marseille, France. European Language Resources Association.

\bibitem[{Wu et~al.(2019)Wu, Yao, Han, Xie, Liu, Lin, Lin, and Sun}]{wu-etal-2019-open}
Ruidong Wu, Yuan Yao, Xu~Han, Ruobing Xie, Zhiyuan Liu, Fen Lin, Leyu Lin, and Maosong Sun. 2019.
\newblock \href {https://doi.org/10.18653/v1/D19-1021} {Open relation extraction: Relational knowledge transfer from supervised data to unsupervised data}.
\newblock In \emph{Proceedings of the 2019 Conference on Empirical Methods in Natural Language Processing and the 9th International Joint Conference on Natural Language Processing (EMNLP-IJCNLP)}, pages 219--228, Hong Kong, China. Association for Computational Linguistics.

\bibitem[{Wu et~al.(2023{\natexlab{a}})Wu, Mangla, Durrett, and Li}]{wu-etal-2023-qudeval}
Yating Wu, Ritika Mangla, Greg Durrett, and Junyi~Jessy Li. 2023{\natexlab{a}}.
\newblock \href {https://doi.org/10.18653/v1/2023.emnlp-main.325} {{QUD}eval: The evaluation of questions under discussion discourse parsing}.
\newblock In \emph{Proceedings of the 2023 Conference on Empirical Methods in Natural Language Processing}, pages 5344--5363, Singapore. Association for Computational Linguistics.

\bibitem[{Wu et~al.(2023{\natexlab{b}})Wu, Sheffield, Mahowald, and Li}]{wu-etal-2023-elaborative}
Yating Wu, William Sheffield, Kyle Mahowald, and Junyi~Jessy Li. 2023{\natexlab{b}}.
\newblock \href {https://doi.org/10.18653/v1/2023.emnlp-main.336} {Elaborative simplification as implicit questions under discussion}.
\newblock In \emph{Proceedings of the 2023 Conference on Empirical Methods in Natural Language Processing}, pages 5525--5537, Singapore. Association for Computational Linguistics.

\bibitem[{Xiao et~al.(2023)Xiao, Liu, Zhang, and Muennighoff}]{bge}
Shitao Xiao, Zheng Liu, Peitian Zhang, and Niklas Muennighoff. 2023.
\newblock \href {http://arxiv.org/abs/2309.07597} {C-pack: Packaged resources to advance general chinese embedding}.

\bibitem[{Xiong et~al.(2023)Xiong, Liu, Molybog, Zhang, Bhargava, Hou, Martin, Rungta, Sankararaman, Oguz, Khabsa, Fang, Mehdad, Narang, Malik, Fan, Bhosale, Edunov, Lewis, Wang, and Ma}]{long-llama}
Wenhan Xiong, Jingyu Liu, Igor Molybog, Hejia Zhang, Prajjwal Bhargava, Rui Hou, Louis Martin, Rashi Rungta, Karthik~Abinav Sankararaman, Barlas Oguz, Madian Khabsa, Han Fang, Yashar Mehdad, Sharan Narang, Kshitiz Malik, Angela Fan, Shruti Bhosale, Sergey Edunov, Mike Lewis, Sinong Wang, and Hao Ma. 2023.
\newblock \href {http://arxiv.org/abs/2309.16039} {Effective long-context scaling of foundation models}.

\bibitem[{Xu and Choi(2022)}]{xu-choi-2022-modeling}
Liyan Xu and Jinho Choi. 2022.
\newblock \href {https://doi.org/10.18653/v1/2022.naacl-main.395} {Modeling task interactions in document-level joint entity and relation extraction}.
\newblock In \emph{Proceedings of the 2022 Conference of the North American Chapter of the Association for Computational Linguistics: Human Language Technologies}, pages 5409--5416, Seattle, United States. Association for Computational Linguistics.

\bibitem[{Xu and Choi(2020)}]{xu-choi-2020-revealing}
Liyan Xu and Jinho~D. Choi. 2020.
\newblock \href {https://doi.org/10.18653/v1/2020.emnlp-main.686} {Revealing the myth of higher-order inference in coreference resolution}.
\newblock In \emph{Proceedings of the 2020 Conference on Empirical Methods in Natural Language Processing (EMNLP)}, pages 8527--8533, Online. Association for Computational Linguistics.

\bibitem[{Xu et~al.(2023{\natexlab{a}})Xu, Zhang, Li, Shang, and Choi}]{xu-etal-2023-towards}
Liyan Xu, Chenwei Zhang, Xian Li, Jingbo Shang, and Jinho~D. Choi. 2023{\natexlab{a}}.
\newblock \href {https://doi.org/10.18653/v1/2023.acl-long.683} {Towards open-world product attribute mining: A lightly-supervised approach}.
\newblock In \emph{Proceedings of the 61st Annual Meeting of the Association for Computational Linguistics (Volume 1: Long Papers)}, pages 12223--12239, Toronto, Canada. Association for Computational Linguistics.

\bibitem[{Xu et~al.(2022{\natexlab{a}})Xu, Zhang, Zong, Liu, Cheng, Ni, Chen, Zhao, and Choi}]{Xu_isdg_2022}
Liyan Xu, Xuchao Zhang, Bo~Zong, Yanchi Liu, Wei Cheng, Jingchao Ni, Haifeng Chen, Liang Zhao, and Jinho~D. Choi. 2022{\natexlab{a}}.
\newblock \href {https://doi.org/10.1609/aaai.v36i10.21407} {Zero-shot cross-lingual machine reading comprehension via inter-sentence dependency graph}.
\newblock \emph{Proceedings of the AAAI Conference on Artificial Intelligence}, 36(10):11538--11546.

\bibitem[{Xu et~al.(2024)Xu, Ping, Wu, McAfee, Zhu, Liu, Subramanian, Bakhturina, Shoeybi, and Catanzaro}]{quality-baseline}
Peng Xu, Wei Ping, Xianchao Wu, Lawrence McAfee, Chen Zhu, Zihan Liu, Sandeep Subramanian, Evelina Bakhturina, Mohammad Shoeybi, and Bryan Catanzaro. 2024.
\newblock \href {https://openreview.net/forum?id=xw5nxFWMlo} {Retrieval meets long context large language models}.
\newblock In \emph{The Twelfth International Conference on Learning Representations}.

\bibitem[{Xu et~al.(2023{\natexlab{b}})Xu, Pang, Li, Yu, Meng, Shen, Cheng, and Zhou}]{xu2023plot}
Shicheng Xu, Liang Pang, Jiangnan Li, Mo~Yu, Fandong Meng, Huawei Shen, Xueqi Cheng, and Jie Zhou. 2023{\natexlab{b}}.
\newblock \href {http://arxiv.org/abs/2311.01666} {Plot retrieval as an assessment of abstract semantic association}.

\bibitem[{Xu et~al.(2022{\natexlab{b}})Xu, Wang, Yu, Ritchie, Yao, Wu, Zhang, Li, Bradford, Sun, Hoang, Sang, Hou, Ma, Yang, Peng, Yu, and Warschauer}]{xu-etal-2022-fantastic}
Ying Xu, Dakuo Wang, Mo~Yu, Daniel Ritchie, Bingsheng Yao, Tongshuang Wu, Zheng Zhang, Toby Li, Nora Bradford, Branda Sun, Tran Hoang, Yisi Sang, Yufang Hou, Xiaojuan Ma, Diyi Yang, Nanyun Peng, Zhou Yu, and Mark Warschauer. 2022{\natexlab{b}}.
\newblock \href {https://doi.org/10.18653/v1/2022.acl-long.34} {Fantastic questions and where to find them: {F}airytale{QA} {--} an authentic dataset for narrative comprehension}.
\newblock In \emph{Proceedings of the 60th Annual Meeting of the Association for Computational Linguistics (Volume 1: Long Papers)}, pages 447--460, Dublin, Ireland. Association for Computational Linguistics.

\bibitem[{Yu et~al.(2023)Yu, Li, Yao, Pang, Zhou, Xiao, Meng, and Zhou}]{yu-etal-2023-personality}
Mo~Yu, Jiangnan Li, Shunyu Yao, Wenjie Pang, Xiaochen Zhou, Zhou Xiao, Fandong Meng, and Jie Zhou. 2023.
\newblock \href {https://doi.org/10.18653/v1/2023.acl-long.826} {Personality understanding of fictional characters during book reading}.
\newblock In \emph{Proceedings of the 61st Annual Meeting of the Association for Computational Linguistics (Volume 1: Long Papers)}, pages 14784--14802, Toronto, Canada. Association for Computational Linguistics.

\bibitem[{Yu et~al.(2024)Yu, Wang, Zhang, Sang, Pu, Wei, Wang, Xu, Li, Yu, and Zhou}]{tominamc}
Mo~Yu, Qiujing Wang, Shunchi Zhang, Yisi Sang, Kangsheng Pu, Zekai Wei, Han Wang, Liyan Xu, Jing Li, Yue Yu, and Jie Zhou. 2024.
\newblock \href {https://openreview.net/forum?id=ZZ7UKgK4c1} {Few-shot character understanding in movies as an assessment to meta-learning of theory-of-mind}.
\newblock In \emph{Forty-first International Conference on Machine Learning}.

\end{thebibliography}

\clearpage

\appendix

\section{Graph Realization}
\label{appx:realization}

\subsection{Full Prompts and Details}
\label{appx:prompts}

Full prompts of our designed two-stage LLM prompting scheme (Section~\ref{sec:graph}) are provided in Figure~\ref{fig:prompt-qg-1}-\ref{fig:prompt-qa}. We specify the maximum number of generated questions for a node pair as 4 in the prompt.

For the task of plot retrieval (Section~\ref{sec:task-rt}) and long document QA (Section~\ref{sec:task-qa}), we construct edges within a neighboring window of 4 preceding nodes, such that the graph realization is proportional to the input instead of being quadratic. For recap identification \cite{recap}, edges are obtained on the provided preceding snippets.

For a context with $T$k tokens, it takes approximately $6T$k tokens to obtain all edge questions of \graph using GPT-4, which costs \$$0.03T$ as of this writing.

\subsection{Qualitative Examples}
\label{appx:examples}

Examples of generated questions on \textit{Game of Thrones} from RECIDENT \cite{recap}.

\subsubsection{Case1}

\textbf{Current Context:}
\begin{adjustwidth}{-0.4cm}{-0.4cm} 
\begin{quote}
\small
\it
At Craster's Keep, Locke scouts the keep for the party of the Night's Watch sent to eliminate the traitors holed up there; in his reconnaissance, Locke finds the hut where Bran Stark, Jojen, Meera and Hodor are being held captive. Reporting back to Jon Snow and the others, Locke tells them that only eleven traitors are present and most of them are drunk and won't prove much of a threat. He also lies about the hut where Jon's brother Bran and his group are being imprisoned, claiming there are only hounds kept inside and that they should keep away from it to prevent the dogs alerting their enemy. Believing Locke, Jon agrees and tells the party they attack at nightfall.
\end{quote}
\end{adjustwidth}

\noindent \textbf{Prior Context:}
\begin{adjustwidth}{-0.4cm}{-0.4cm} 
\begin{quote}
\small
\it
Having received word of the wildlings' raids down south, the Lord Commander states that they do not have the manpower to afford venturing away from the Wall. They are interrupted when Edd and Grenn return to Castle Black after escaping Craster's Keep. Jon reveals he told Mance Rayder that a thousand men armed Castle Black and therefore points out that when Mance reaches Craster's Keep, Rast and Karl Tanner will not hesitate in revealing the truth. Jon then insists the Night's Watch send a party to Craster's Keep to kill their traitor brothers before Mance gets to them first.
\end{quote}
\end{adjustwidth}

\noindent \textbf{Generated Question (Valid):}
\begin{adjustwidth}{-0.4cm}{-0.4cm} 
\begin{quote}
\small
\sl What prompted the Night's Watch to act with urgency in sending a party to Craster's Keep to eliminate the traitors?
\end{quote}
\end{adjustwidth}

\noindent \textbf{Generated Question (Invalid):}
\begin{adjustwidth}{-0.4cm}{-0.4cm} 
\begin{quote}
\small
\textsl{What was the reason behind Jon Snow's insistence on a strategic assault to silence the traitors before a specific event could occur?}

(Note: it is a question asked upon the prior context and can be answered by it directly, as addressed in the Limitations Section, not bridging two context.)
\end{quote}
\end{adjustwidth}

\subsubsection{Case2}
\textbf{Current Context:}
\begin{adjustwidth}{-0.4cm}{-0.4cm} 
\begin{quote}
\small
\it
In what becomes known as the infamous Red Wedding, Lothar draws a knife and repeatedly stabs the pregnant Talisa in the stomach, killing her unborn child. Talisa collapses to the ground as chaos surrounds. Before he can react, Robb is shot by the musicians with crossbows several times and falls to the floor. Numerous other Stark men are killed by the crossbow bolts or set upon by Frey soldiers. Catelyn is shot by one of the musicians in the back and falls to the floor.
\end{quote}
\end{adjustwidth}

\noindent \textbf{Prior Context:}
\begin{adjustwidth}{-0.4cm}{-0.4cm} 
\begin{quote}
\small
\it
In Gendry's quarters, Melisandre seduces Gendry long enough to distract him, then promptly ties him to the bed and places leeches on his body. She explains as Stannis and Davos enter the room that Davos wanted a demonstration of the power in king's blood, then removes the leeches and lights a fire in a nearby brazier. As part of the magical ritual that follows, Stannis throws the leeches into the flames at Melisandre's direction, and recites the names of three people he wants dead as they burn: "The usurper Robb Stark, the usurper Balon Greyjoy, the usurper Joffrey Baratheon."
\end{quote}
\end{adjustwidth}

\noindent \textbf{Generated Question (Valid):}
\begin{adjustwidth}{-0.4cm}{-0.4cm} 
\begin{quote}
\small
\textsl{What ritual was performed prior to the Red Wedding that sought the death of Robb Stark and might have influenced his fate?}

(Note: it is an open question whether Melisandre's ritual really worked and is widely discussed among fans. The question uses \emph{might} which adds its accuracy.)

\end{quote}
\end{adjustwidth}

\subsubsection{Case3}
\textbf{Current Context:}
\begin{adjustwidth}{-0.4cm}{-0.4cm} 
\begin{quote}
\small
\it
In King's Landing, Eddard is summoned to the throne room by "King Joffrey"; Robert has died. He arrives to find Littlefinger and Varys waiting for him, along with Commander Janos Slynt and a detachment of the City Watch. Varys tells him that Renly has fled the city, along with Ser Loras Tyrell and a number of retainers. They were last seen heading south. The party enters the throne room, where Joffrey sits on the Iron Throne. He demands oaths of fealty from his councilors and subjects. Instead, Eddard gives Ser Barristan Selmy the proclamation naming him as Lord Protector of the Realm. To Barristan's shock, Cersei takes the "paper shield" and tears it up. Instead, she suggests that Eddard bend the knee and swear allegiance.
\end{quote}
\end{adjustwidth}

\noindent \textbf{Prior Context:}
\begin{adjustwidth}{-0.4cm}{-0.4cm} 
\begin{quote}
\small
\it
Lord Eddard Stark meets with Cersei Lannister. He tells her that he knows the secret that Jon Arryn died for: that Cersei's three children are not Robert's, but the product of incest between her and Jaime. Cersei does not deny the charge and in fact is proud of it, comparing their love to the old Targaryen practice of marrying brother to sister; she also admits to having despised Robert ever since their wedding night, when Robert drunkenly stumbled into Cersei's bed and called her "Lyanna". Eddard angrily tells her to take her children and leave the city immediately. When Robert returns from his hunt, he will tell him the truth of the matter and Cersei should run as far as she can before that happens, lest Robert's wrath find her.
\end{quote}
\end{adjustwidth}

\noindent \textbf{Generated Question (Valid):}
\begin{adjustwidth}{-0.4cm}{-0.4cm} 
\begin{quote}
\small
\sl
What is the reason behind Eddard Stark's refusal to swear fealty to Joffrey and his decision to present a proclamation in the throne room?
\end{quote}
\end{adjustwidth}

\noindent \textbf{Generated Question (Invalid):}
\begin{adjustwidth}{-0.4cm}{-0.4cm} 
\begin{quote}
\small
\textsl{What prevented Eddard Stark from informing King Robert about the illegitimacy of Cersei's children, which could have significantly altered the succession to the Iron Throne?}
\end{quote}
\end{adjustwidth}

\subsubsection{Case4}
\textbf{Current Context:}
\begin{adjustwidth}{-0.4cm}{-0.4cm} 
\begin{quote}
\small
\it
In what becomes known as the infamous Red Wedding, Lothar draws a knife and repeatedly stabs the pregnant Talisa in the stomach, killing her unborn child. Talisa collapses to the ground as chaos surrounds. Before he can react, Robb is shot by the musicians with crossbows several times and falls to the floor. Numerous other Stark men are killed by the crossbow bolts or set upon by Frey soldiers. Catelyn is shot by one of the musicians in the back and falls to the floor.
\end{quote}
\end{adjustwidth}

\noindent \textbf{Prior Context:}
\begin{adjustwidth}{-0.4cm}{-0.4cm} 
\begin{quote}
\small
\it
At Harrenhal, Jaime speaks one last time to Brienne before he leaves. Jaime remarks that he owes Brienne a debt for both keeping him alive on their journey and for giving him a reason to live to rouse him from his suicidal depression after losing his hand. Brienne tells Jaime to repay his debt by keeping his pledge. Jaime promises that he will keep his word to return Catelyn Stark's daughters to her.
\end{quote}
\end{adjustwidth}

\noindent \textbf{Generated Question (Invalid):}
\begin{adjustwidth}{-0.4cm}{-0.4cm} 
\begin{quote}
\small
\textsl{What prior commitment made by Jaime Lannister could influence the fate of the Stark family following the Red Wedding, where Catelyn Stark is among those attacked?}

(Note: the question is rather irrelevant in regards to 
 the two context snippets.)
\end{quote}
\end{adjustwidth}

\subsubsection{Case5}
\textbf{Current Context:}
\begin{adjustwidth}{-0.4cm}{-0.4cm} 
\begin{quote}
\small
\it
Tormund and Beric Dondarrion review the defenses atop the Wall at Eastwatch-by-the-Sea. Tormund remarks that the crows say he'll get used to the height, but he admits it'll probably be a while. Suddenly, the pair sees movement at the edge of the Haunted Forest. A White Walker emerges atop an undead horse, followed shortly by a horde of wights. More and more White Walkers emerge as the Night Watch's horns sound three times. However, the army of the dead stops some distance from the foot of the Wall and Tormund looks relieved; despite their numbers, the dead don't have anything that could possibly get them past the barrier. But then all on the Wall stop in horror as they hear a very familiar sound; a screeching roar mixed with the heavy thumping of huge wings beating the air.
\end{quote}
\end{adjustwidth}

\noindent \textbf{Prior Context:}
\begin{adjustwidth}{-0.4cm}{-0.4cm} 
\begin{quote}
\small
\it
At Eastwatch, Sandor carries the struggling Wight into a boat. Tormund and Beric tell him they will meet again but Sandor retorts he hopes not. Daenerys sends Drogon and Rhaegal to scour the surrounding mountains for Jon. Jorah tells Daenerys that it is time to leave but she insists on waiting a bit longer. Before she can leave, they hear a horn blowing signaling a rider approaching. Looking down from the battlements, Dany sees a wounded Jon Snow approaching on horseback. Aboard their ship, Davos and Gendry remove the frozen-stiff garments and tend to Jon Snow, who has suffered severe hypothermia and several minor injuries. Daenerys also notes the massive scars on his chest from his previous fatal wounds.
\end{quote}
\end{adjustwidth}

\noindent \textbf{Geneated Question (Invalid):}
\begin{adjustwidth}{-0.4cm}{-0.4cm} 
\begin{quote}
\small
\textsl{What was Daenerys waiting for at Eastwatch before Jon Snow's wounded arrival on horseback?}

(Note: this is another example of asking upon the prior context, which could happen more often than irrelevant questions.)
\end{quote}
\end{adjustwidth}

\subsection{Experiments}
\label{appx:exp}

\paragraph{LLM}
The usage of ChatGPT (\textit{gpt-3.5-turbo}) and GPT-4 (\textit{gpt-4-1106-preview}) is through OpenAI's paid API service. For the open-source Llama-2 \cite{llama2}, we perform inference on Nvidia A100 GPUs.

\paragraph{Training}
For training a rerank model in Section~\ref{sec:task-rt}, we initialize a BERT model with weights from BGE-Large \cite{bge}, and use the mean-pooled token embeddings as the sequence representation, following the standard S-BERT setup \cite{reimers-gurevych-2019-sentence}.
The training is conducted on one Nvidia A100 GPU, taking around 6 hours to finish, with 20 epochs, 20 queries within each batch, learning rate $2\times10^{-5}$, cosine learning rate schedule, and a warmup ratio of $5\times10^{-2}$.

\clearpage


\begin{figure*}[ht!]
    \centering
\begin{lstlisting}[language=prompt]

You are an expert on reading and analyzing a wide variety of books. Given the following two snippets [[[snippet_a]]] and [[[snippet_b]]] from a book, where [[[snippet_a]]] happens before [[[snippet_b]]], you need to find concrete parts in both snippets that reflect this temporal relation, such that certain parts in [[[snippet_a]]] contribute as the preceding background or cause for specific events or situations in [[[snippet_b]]].

|||[snippet_a]|||

|||[snippet_b]|||

Please try your best to provide a brief markdown list of each important point that contains those specific parts from both snippets and briefly explains how one serves as the background or cause for the other so to reflect their temporal or causal relation (no more than four points in total). 
Note that only list evident and important points without much guessing; it is ok to find only one, or even no such points.
\end{lstlisting}
    \caption{Prompt for Question Generation (turn 1). Slots in \textcolor{blue}{blue} refer to the input texts.}
    \label{fig:prompt-qg-1}
\end{figure*}


\begin{figure*}[ht!]
    \centering
\begin{lstlisting}[language=prompt]
Please convert each of your listed point to the form of question, such that each question asks about the cause or background (rather than outcome or consequence) of specific events or situations mentioned in [[[snippet_b]]], which can be answered or clarified by the corresponding part in the preceding [[[snippet_a]]]. Hence, these questions should be helpful to reflect their temporal or causal or other important relations between the two snippets. Note that the question should ask upon specific things from [[[snippet_b]]] that cannot be answered by [[[snippet_b]]] itself, and should be answerable by concrete parts from [[[snippet_a]]] without disclosing those parts directly in the question.

Please try your best to think of one such question for each listed point; for your response, return each question starting with "Q:". 
Questions should be asked directly without mentioning "snippet" or any other explanation; questions should be concise but also provide necessary context to avoid ambiguity.

\end{lstlisting}
    \caption{Prompt for Question Generation (turn 2). Slots in \textcolor{blue}{blue} refer to the input texts.}
    \label{fig:prompt-qg-2}
\end{figure*}


\begin{figure*}[ht!]
    \centering
\begin{lstlisting}[language=prompt]
You are an expert on reading and analyzing a wide variety of books. Given the following snippet [[[snippet]]] from a book, and a related question
 [[[question]]], you need to determine whether the provided snippet could answer this question.

|||[snippet]|||

|||[question]|||

Please first reason the question very briefly, then give the judgement. If the provided snippet does not present useful information to answer the question, print [UNANSWERABLE] after the reasoning and terminate your response. Otherwise, if the question is indeed answerable, print [ANSWERABLE] after the reasoning, immediately followed by a concise markdown list of the most crucial original sentences from the snippet that could serve as the key supporting evidence for the answer of the question; directly show each sentence per line, without any extra explanation.
\end{lstlisting}
    \caption{Prompt for Question Filtering via back verification. Slots in \textcolor{blue}{blue} refer to the input texts.}
    \label{fig:prompt-qa}
\end{figure*}

\end{document}